\documentclass[11pt,letterpaper]{article}

\usepackage[T1]{fontenc}

\usepackage{cleveref}
\usepackage[subject={Todo}]{pdfcomment}

\DeclareRobustCommand{\fnurl}[1]{\footnote{\url{#1}}}

\usepackage{amsmath}
\usepackage{booktabs} % For formal tables
\usepackage{times}
\usepackage{url}
\usepackage{wrapfig}
\usepackage{bm}
\usepackage{xcolor,colortbl}
\usepackage{multirow}
\usepackage[para]{threeparttable}
\usepackage{listings}
\usepackage{csquotes}
\usepackage{graphicx}
\usepackage{tabularx} % better tables10,twocolumn,latex8]{article}
\usepackage{comment}
\usepackage{enumitem}
\setenumerate{noitemsep}
\setitemize{noitemsep}
\setdescription{noitemsep}
\usepackage{subfig}

\usepackage{tikz}
\usepackage{lipsum}

\usepackage{geometry}
\usepackage[utf8]{inputenc}

\usepackage{cite}

\usepackage{nameref,hyperref}

\usepackage[right]{lineno}

\usepackage{microtype}
\DisableLigatures[f]{encoding = *, family = * }

\usepackage{changepage}

\usepackage[justification=centering]{caption}

\usepackage{lastpage,fancyhdr,graphicx}
\usepackage{epstopdf}
\fancyheadoffset[L]{2.25in}
\fancyfootoffset[L]{2.25in}

\usepackage{color}

\definecolor{Gray}{gray}{.25}

\usepackage{graphicx}

\usepackage{sidecap}

\usepackage{wrapfig}
\usepackage[pscoord]{eso-pic}
\begin{document}
\vspace*{0.35in}

% title goes here:
\begin{flushleft}
{\Large
\textbf{Collaborations on YouTube: From Unsupervised Detection to the Impact on Video and Channel Popularity}
}
\\
% authors go here:
Christian Koch,
Moritz Lode,
Denny Stohr,
Amr Rizk,
Ralf Steinmetz
\\
Multimedia Communications Lab (KOM), Technische Universit\"at Darmstadt, Germany
\\
E-Mail: \{Christian.Koch | Denny.Stohr | Amr.Rizk | Ralf.Steinmetz\}@kom.tu-darmstadt.de 
\end{flushleft}
%\blfootnote{© 2018, The copyright remains with the authors.}

\section*{Abstract}
YouTube is one of the most popular platforms for streaming of user-generated video.
Nowadays, professional YouTubers are organized in so called multi-channel networks (MCNs).
These networks offer services such as brand deals, equipment, and strategic advice in exchange for a share of the YouTubers' revenue.
A major strategy to gain more subscribers and, hence, revenue is collaborating with other YouTubers.
Yet, collaborations on YouTube have not been studied in a detailed quantitative manner.
This paper aims to close this gap with the following contributions.
First, we collect a YouTube dataset covering video statistics over three months for 7,942 channels.
Second, we design a framework for collaboration detection given a previously unknown number of persons featuring in YouTube videos. We denote this framework for the analysis of collaborations in YouTube videos using a Deep Neural Network (DNN) based approach as CATANA.
Third, we analyze about 2.4 years of video content and use CATANA to answer research questions providing guidance for YouTubers and MCNs for efficient collaboration strategies.
Thereby, we focus on \emph{(i)} collaboration frequency and partner selectivity, \emph{(ii)} the influence of MCNs on channel collaborations, \emph{(iii)} collaborating channel types, and \emph{(iv)} the impact of collaborations on video and channel popularity.
Our results show that collaborations are in many cases significantly beneficial in terms of viewers and newly attracted subscribers for both collaborating channels, showing often more than 100\% popularity growth compared with non-collaboration videos.

\vspace{0.25cm}

% now start line numbers
%\linenumbers

\section{Introduction \& Problem Statement} \label{intro}
%Problem Statement
Collaborations are a key strategy to increase the dissemination and, hence, popularity of user-generated videos on YouTube.
%Collaborations are a key strategy to increase the dissemination of user created content on YouTube.
%increase the popularity of created content and hence to reach a broader audience.
%, e.g., research papers,  industry projects, as well as user-generated content (UGC) such as YouTube videos.
Here, YouTuber \emph{A}, i.e., a content creator, introduces a collaborating YouTuber \emph{B} to his viewers seeking to attract their interest to the content of YouTuber \emph{B}.
The rationale here is that given the viewers interests, they are more likely to watch the introduced YouTuber's videos and eventually become subscribers~\cite{Gahan2015,TubeFilter2015,CreatorAcademy2016}.
Collaborations often occur in a reciprocal manner on both of the collaborating channels to attract each other's viewers and, thereby, increase both YouTuber's revenue.
%A key These viewers are likely to become subscribers, i.e., regular viewers of his videos as well~\cite{Gahan2015,TubeFilter2015,CreatorAcademy2016}.
Video monetization can be activated through participating in the YouTube Partner Program (YPP)~\cite{Holmbom2015}.
Thereby, YouTubers can configure their videos allowing YouTube to insert short ad clips before and within their videos.
In return, YouTube pays a share of the resulting revenues out to the YouTubers.
With more than 10k people that were reported in 2016, the annual growth of the number of YouTubers earning six figures per year amounts to $\sim50\%$~\cite{Ytpress2016}.

In contrast to works describing collaborations, e.g., within scientific communities or online social networks, there are only few studies on the detection of collaborations in
user-generated videos and a lack of quantitative analysis of their desired impact on popularity.
Note that Video-on-Demand (VoD) platforms such as Netflix are not exposed to user-generated content (UGC) which is more scattered, diverse, and generated at a much higher rate.
While general metadata of UGC videos is available, no explicit information on collaborations exists.
Although mentions of collaborating YouTubers in the video title or in the description are in general possible, they often miss collaboration-related information and are, hence, not a reliable source of information.

% MCN perspective
Nowadays, popular YouTubers often belong to a multi-channel network (MCN), e.g., BroadbandTV, Studio71, and Maker Studios, which offer equipment, brand deals, and strategic advice to increase the YouTubers' popularity in exchange for a share of their revenue~\cite{Holmbom2015}.
Essentially, such strategic advice comprises collaboration policies that describe the most beneficial collaborations with diverse YouTubers.

Until now, a public documentation on the efficiency of collaborations on YouTube does not exist.
A quantitative analysis of different types of collaborations, e.g., between YouTubers of different popularity or content categories has the potential to guide YouTubers to popularity- and, hence, revenue-maximizing behavior.
To this end, this article addresses the challenging task of identifying and analyzing content creator collaborations on large-scale UGC video platforms taking the example of YouTube.
Our analysis focuses on \emph{(i)} collaboration frequency and partner selectivity, \emph{(ii)} the influence of MCNs on channel collaborations, \emph{(iii)} collaborating channel types, and \emph{(iv)} the impact of collaborations on video and channel popularity.
In the following sections we address these focus points in a framework for \textbf{c}ollabor\textbf{at}ion detection and \textbf{ana}lysis denoted CATANA.
For the design of CATANA, we argue that the analysis of content creator collaborations on large-scale UGC video platforms faces three key challenges:
%is an interesting but also challenging research area.
%
%We identify three key challenges:

\begin{itemize}

\item \textbf{YouTube-suitable Face Detection:} The YouTubers' faces constitute the basis for identifying collaborations.
Hence, we require their inference and storage in an appropriate representation.
Changing lights, face expressions, and camera perspectives add to the difficulty of deriving these representations.
 %need for each face need to result in a similar representation to determine if it is the same persons.

\vspace{0.15cm}

\item \textbf{YouTuber Identification:} In contrast to most face recognition tasks, in our scenario, an unknown number of people appearing in user-generated videos needs to be identified based on the detected faces.
Therefore, an accurate association of face samples to people for a large set of YouTubers needs to be obtained.
Additionally, we need to identify which person is the owner of the YouTube channel and which persons are potential guests.

\vspace{0.15cm}

\item \textbf{Collaboration Representation:} The co-occurrence of YouTubers in a video indicates collaboration.
However, people appearing sporadically, e.g., passersby do not carry strong evidence of collaborations.
Filtering outliers and storing inferred collaborations in an appropriate form is essential for further analysis.

\end{itemize}

% 2 Contributions: unsupervised collaboration detection and characterizing the impact of collaborations on the video popularity

%We address the first challenge of accurate face detection for UGC by benchmarking the performance of seven approaches including deep learning-based ones on two recent YouTube datasets showing that
% in \cref{design_face_detection_and_recognition}.
%Facenet, the method of choice, achieves an accuracy of $\geq 99.3\%$ for both datasets.
% and is, hence, the best performing approach.

\vspace{0.15cm}

The remainder of this article is structured as follows.
\Cref{related_work} provides an overview of used benchmarking datasets, face recognition approaches, and existing works on YouTube collaborations.
In \Cref{system_design}, we motivate and explain CATANA's major building blocks.
\Cref{data_aquisition} presents our data acquisition and selection process.
In \Cref{eval}, we evaluate and discuss our results answering relevant research questions.
We conclude our article in \Cref{conclusion} and discuss future work.
\section{Related Work}\label{related_work}
%=========================================
%Not much work has been done in the area of collaboration analysis on content portals like YouTube so far.

The related work provides, first, an overview of existing datasets that are useful for benchmarking face recognition methods on YouTube.
Second, an overview of recent face recognition methods is given.
Third, existing works addressing YouTube collaboration analysis are presented and discussed.

\subsection{YouTube Face Recognition Datasets}
In this section we give a brief overview of the two face datasets used for optimizing CATANA.
Both datasets are frequently used in face recognition studies of the last years, as well as in related work.

\paragraph{Labeled Faces in the Wild (LFW)} LFW is a publicly available face dataset introduced in~\cite{LFWTech} and released as an effort to spur face recognition research.
It has been referenced in more than 50 papers related to face recognition~\cite{LFW2015}.
LFW's provides a large set of face images in a large range of variations.
This includes variation in pose, lighting, expression, background, ethnicity, and age~\cite{LFW}.
The dataset consists of 13,244 images of 5,749 people with images in 250x250 pixel JPEG format.
These face images were collected from the web and contain celebrities, politicians, athletes, and other public figures.
For training, validation, and testing, mutually exclusive splits are available suited for usage in a 10-fold cross validation.
Training and testing data are, thereby, presented as either matching or mismatching face pairs.
Hence, the proposed evaluation experiments aim at the problem of pair matching, deciding whether the images are of the same person.
In total 6,000 pairs are provided, divided into 10 splits with each 300 match and 300 mismatch pairs.

\paragraph{YouTube Faces (YTF)}\noindent
YTF is a dataset of face videos introduced in~\cite{YTFPaper} and designed for studying the problem of face recognition in videos.
The dataset contains 3,425 videos of 1,595 different people appearing in YouTube videos.
These people are a subset of the 5,749 people from the LFW dataset and for every person an average of 2.15 videos are available.
The video duration ranges between 48 and 6,070 frames and all video frames are stored individually as JPEG images.
Structure and design of YTF is heavily inspired by LFW.
Similarly, matching and mismatching pair are provided, for usage in a 10-fold cross validation pair-matching evaluation.
In comparison to the LFW evaluation, pairs consist of videos, allowing to decide if the pair of videos is subject of the same person or not.
Overall, 5,000 pairs are provided, divided into 10 splits, each containing 250 match and 250 mismatched pairs.

\subsection{Face Recognition}\label{related-work-face-recognition}
Two kinds of fundamentally different face recognition approaches exist.
The first kind that is widely-spread relies on fixed mathematical models and structures, e.g.,
Eigenfaces\cite{Turk1991} and Local Binary Patterns Histograms (LBPH) \cite{Ahonen2006}.
While these classical approaches have shown solid performance, the second kind of techniques, using deep neural networks (DNNs) is outperforming classical approaches and beginning to replace them.
As they are clearly superior in performance to classical approaches and even compared with humans~\cite{LFW2015}, we focus only on DNN-based approaches.

Parkhi et al.~\cite{Parkhi2015} present a deep learning-based face recognition method called VGG-Face. The authors further propose how a very large-scale face dataset can be assembled by a combination of automation and human in the loop.
Focusing on celebrities for the dataset acquisition, they extract a small number of images using the Google Image Search and the actors' names.
Then, these images are audited through human annotators.
In the next step, the actors are queried again through image search, this time extracting a bigger amount of images.
Instead of human annotation, a linear Support Vector Machine (SVM) is pre-trained with the small number of images from the first step and leveraged to classify the newly acquired images.
Thereby, a dataset of 2,622 identities and 2 million images is assembled.
Furthermore, a CNN (Convolutional Neural Network) based face recognition method is proposed, which is trained on the acquired dataset using the \emph{triplet-loss function}.
The proposed method is then evaluated on the LFW and YTF datasets and compared against other state-of-the-art methods, resulting in an accuracy of 98.95\% for LFW and 97.3\% for YTF.
We conclude that image acquisition through online image search can be a promising method to acquire training data.
In the context of this work, channel names could be queried, resulting in face images of YouTubers.
However, this would still require a human interaction for a potentially large number of YouTubers and passersby in videos.

Schroff et al.~\cite{Schroff2015} present a DNN model named FaceNet.
They introduce the \emph{triplet loss} for training, i.e., a loss function designed to result in face representations, clustered based on similarity.
Triplet loss thereby works with triplets of matching / non-matching face images and embeds these into a 128-dimensional Euclidean space.
Thereby, the loss function ensures that an anchor image of a specific person is closer to all other positive images of the same person,
than it is to any negative image from any other person.
Thus, the network results in a 128-dimensional face representation in a space where distances directly correspond to face similarity.
Consequently, classification and clustering are now straightforward by using standard Euclidean distance metrics.
Hence, accuracies of 99.63\% on the LFW dataset and 95.1\% on YTF were measured.
Even though similar in architecture than the previous DNN~\cite{Parkhi2015}, a significantly larger private dataset of 200M images was used for training.

Amos et al.~\cite{Amos2016} present OpenFace\footnote{\url{http://cmusatyalab.github.io/openface} [Accessed: \today]}, a CNN-based face detection and recognition framework using only openly available training datasets for training.
OpenFace builds up on the FaceNet~\cite{Schroff2015} architecture, combining it with computer vision and machine learning techniques, such as State Vector Machine (SVM) for classification.
For training the CNN model, public face datasets were used, which licenses allow usage and publication of the resulting models.
Hence, no training is necessary for using the OpenFace framework.
OpenFace is thereby trained with only 500,000 images from combining the two largest labeled face recognition datasets for research, CASIA-WebFace\footnote{\url{http://www.cbsr.ia.ac.cn/english/CASIA-WebFace-Database.html} [Accessed: \today]} and FaceScrub\footnote{\url{http://vintage.winklerbros.net/facescrub.html} [Accessed: \today]}.
Overall, OpenFace achieves an accuracy of 92.92\% on LFW.
%OpenFace is made available under the Apache 2.0 license.

A related framework to OpenFace is Facenet\footnote{\url{https://github.com/davidsandberg/facenet} [Accessed: \today]}, as both are based on FaceNet~\cite{Schroff2015} and its proposed DNN.
Even though Facenet shares almost the same name as FaceNet~\cite{Schroff2015}, its developer have no connection to the authors.
In comparison to OpenFace, Facenet additionally implements multiple additional ideas from different papers to tune accuracy.
Amongst others, a different loss function is used for training, the face representation is 1792-dimensional instead of 128-dimensional and, furthermore, a different face detection technique is used~\cite{MTCNN}.
Facenet implements the \emph{center loss} function~\cite{Wen2016}, which is developed to improve the discriminative power of the learned features, minimizing the intra-class variation while keeping the classes separable.
To do so, it learns a center for each feature class and penalizes distances between features and their center, resulting in high accuracy.
Facenet is trained using the MS-Celeb-1M face dataset~\cite{guo2016ms},
whose license allows model reproducibility.
Facenet is evaluated on the LFW dataset, showing an accuracy of 99.2\%.

We conclude that DNN-based approaches show high accuracy on relevant datasets such as LFW and YTF.
As Facenet evidently outperforms the other reviewed approaches, we will deploy it as a key component for CATANA's face recognition capabilities.

\subsection{Collaborations on YouTube}
Mattias Holmbom~\cite{Holmbom2015} conducted a study on five YouTube channels, including interviews with their content creators. He found that it is currently more difficult than ever for new content creators to establish a YouTube channel, as already a large number of channels exist.
Concerning collaborations on YouTube, three of the five YouTubers associate their popularity directly or indirectly with collaborations featuring other channels.
Furthermore, MCNs were suggested as tool to get help in finding other YouTubers addressing similar topics for collaboration.

In Brendan Gahan's article "How to be successful on YouTube: The 3 steps"~\cite{Gahan2015}, the third proposed step for new content creators is collaboration with other YouTubers, which already have an established subscriber base as an essential step to expand a channel's audience.
YouTube's official Creator Academy\footnote{YouTube website offering free online courses helping users creating better videos and improve channel performance.}
also suggests collaboration as a powerful way to reach new users~\cite{CreatorAcademy2016}.
This is also confirmed from MCNs' side, as the MCN "Channel Frederator Network" recommends to facilitate and instigate collaborations between network members~\cite{TubeFilter2015}.
We conclude that analyzing channels registered in the same network can significantly improve the likelihood detecting collaborations.
This is important for our work to control the comparisons of videos with and without collaborations.

Bertram Gugel created a visualization of a share of the YouTube collaboration graph~\cite{Gugel2015}.
The visualized graph provides insights into (i) the size of the respective YouTube channels, measured by subscriptions, (ii) whether the association is uni- or bidirectional, as well as, (iii) to which MCN the channels belong.
However, only the \emph{Featured Channel List} of a YouTube channel is used to determine a collaboration.
On the one hand, this deliberate association between YouTube channels does not necessarily indicate that channels collaborate.
On the other hand, there exists also collaborations with not featured channels.
Furthermore, this work does not provide a thorough analysis of individual videos and the effect of collaboration on their popularity.
Hence, we conclude that to the best of our knowledge no comprehensive analysis on YouTube collaboration and resulting effects on popularity exists so far. We set this as the goal of this article.

%================================
\section{System Design}\label{system_design}
\begin{figure}[t]
	\centering
	\includegraphics[width=0.65\linewidth]{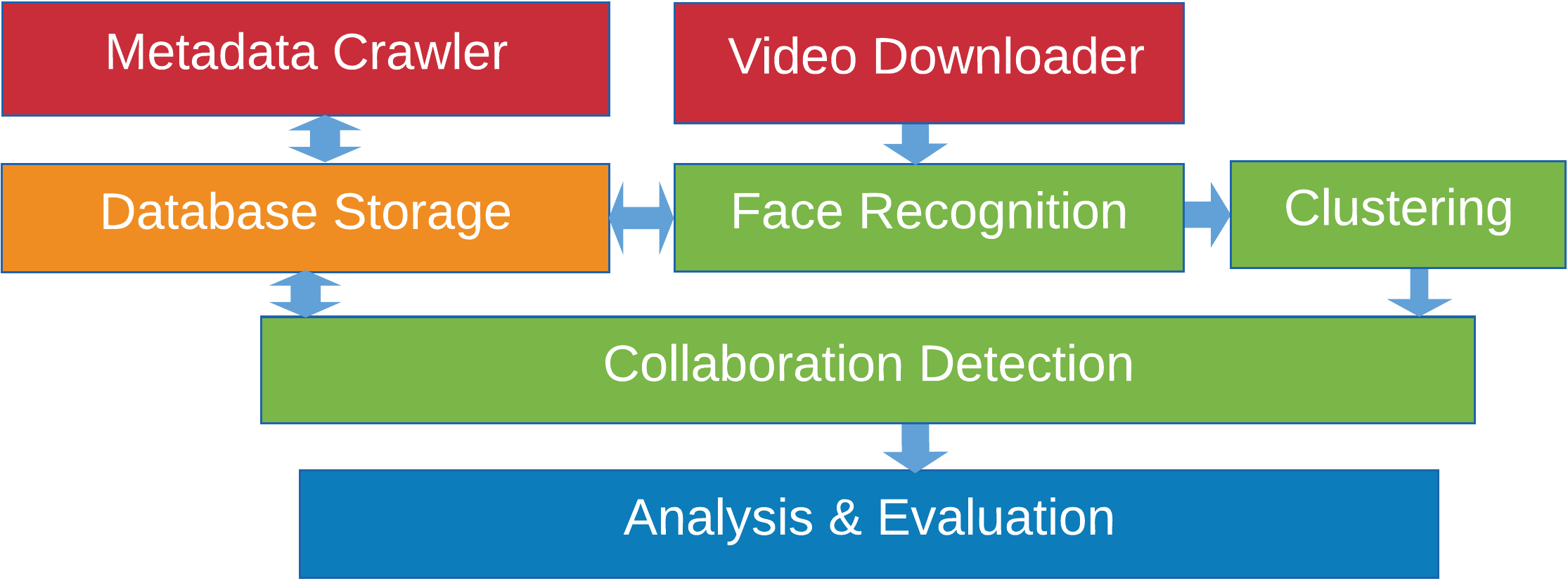}
	\caption{CATANA's system architecture.}
	\label{fig:system-arch}
\end{figure}

%================================
In the related work, we have seen that it is common practice to recommend YouTubers to collaborate to increase their audience and, hence, revenue.
However, there is a lack of precise and detailed analysis of the impact of YouTube collaborations on video and channel popularity.
With CATANA, we aim to close this gap.
In the following, we motivate CATANA's major building blocks, depicted in \Cref{fig:system-arch}.
As our focus is on video and channel popularity, we need the \emph{Metadata Crawler} which collects popularity time series, i.e., information of view counts in the case of videos and of the subscriber counts in the case of channels.
These time series are stored in the \emph{Database Storage}.
Before analyzing the videos, the \emph{Video Downloader} acquires the video data.
In a next step, the contained faces of the videos are detected and a dense representation is computed and stored in the \emph{Database Storage}, together with the video ID allowing to associate the contained faces with the videos' popularity time series.
Then, the \emph{Clustering} module determines face-person associations by associating similar face representations with the same person.
The most representative face representations per person are stored in the \emph{Database Storage}.
The \emph{Collaboration Detection} module compares face representations of different videos, thereby allowing to find persons appearing in different videos, as these videos contain the same face representations, i.e., the same persons.
In a last step, the collaborations detected are stored in form of a bidirectional graph used in the  \emph{Analysis \& Evaluation} module.
In the following, we discuss the key processes and design choices of CATANA in detail.

\subsection{Face Detection and Recognition} \label{design_face_detection_and_recognition}
We evaluated six recent approaches (ref. \Cref{related-work-face-recognition}) to choose an appropriate face recognition method.
Eigenfaces\cite{Turk1991} uses principal component analysis (PCA) to transform a high-dimensional face image to a lower-dimensional representation, while Local Binary Patterns Histograms (LBPH) \cite{Ahonen2006}  are histogram-based.
In contrast to these traditional well-known methods, we also evaluate recent DNN-based approaches: Facenet, Openface, FaceNet, and VGG.
We compare the performance of these approaches with two benchmark datasets: LFW and YTF (ref. \Cref{related-work-face-recognition}).
%Labeled Faces in the Wild~\cite{LFWTech} (LFW) provides a large set of face images of 13,244 images from 5,749 people, in a large range of variations including pose, lighting, expression, background, ethnicity, and age.
%YouTube Faces~\cite{YTFPaper} (YTF) is designed for the evaluation of face recognition in videos.
%It contains 3,425 videos of 1,595 people.
%The contained faces are a subset of the faces from the LFW dataset.
Both datasets provide pairs of matching and non-matching faces useful to assess face recognition methods.
The results are presented in \Cref{tab:dg_face_recog_compare}, depicting the accuracy of the pairwise match/mismatch between subject pairs in a 10-fold cross-validation.
%Both, Facenet and FaceNet show the highest performance on the LFW dataset.
%For YTF, Facenet and OpenFace perform best.
As Facenet performs best on both datasets, and even outperforms humans, we choose it for CATANA's face detection and face recognition functionalities.

\begin{table}[t]
	\caption{Face recognition performance comparison with stDev, non-public are grayed out. \label{tab:dg_face_recog_compare}}
	\small
	\centering
	\begin{threeparttable}
		\begin{tabular}{l l l}
			\toprule
			& \textbf{LFW}       & \textbf{YTF}      \\ \midrule
			\textbf{Eigenfaces}          & 0.6002 $\pm$ 0.0079\tnote{1}   & -                 \\
			\textbf{LBPH}                & 0.6782 $\pm$ 0.6300   & -                 \\
			\textbf{Facenet}             & 0.9930 $\pm$ 0.004\tnote{2}     & 0.9980 $\pm$ 0.0013\tnote{4} \\
			\textbf{Openface}            & 0.9292 $\pm$ 0.0134\tnote{3}    & 0.9971 $\pm$ 0.0026\tnote{4} \\
			\textbf{Human}               & 0.9753\tnote{1}             & -                 \\
			\rowcolor[HTML]{C0C0C0}
			\textbf{FaceNet}            & 0.9963 $\pm$ 0.0009 & 0.9512 $\pm$ 0.39     \\
			\rowcolor[HTML]{C0C0C0}
			\textbf{VGG}                & 0.9913          & 0.9740            \\ \bottomrule
		\end{tabular}
		\begin{tablenotes}\footnotesize
			\item[1] ~\cite{LFW}
			\item[2] ~\cite{Amos2016}
			\item[3] ~\cite{FacenetGH}
			\item[4] this article
		\end{tablenotes}
	\end{threeparttable}
\end{table}

\subsection*{Frame Extraction and Selection}
Typically, face recognition approaches are designed for images, not for video.
Due to the large number of frames per second  with usually small changes, it is not efficient to process every frame.
Instead, we use a duration-based frame extraction rate $f(n,r)$ taking the number of frames $n$ in the video and its frame rate $r$ as inputs to reduce the number of images to process. 
We extract evenly spaced frames with a usual rate of 10 frames per minute. 
For short video, we adapt the rate to extract at least $f_{\text{min}}=600$  frames which shows a good performance. 
For long videos, the number of frames extracted is limited for storage reasons to $f_{\text{max}}$=8,000.
%
%%, see \Cref{eq:df_frame_extraction_eq}.
%%
%\begin{equation}
%	f(n, r)=\begin{cases}
%		(\frac{1}{6} \times \frac{n}{r})+f_{min},& \text{if } (\frac{1}{6} \times \frac{n}{r})+f_{min} < f_{max}.\\
%		f_{\text{max}},             & \text{otherwise}.
%	\end{cases}
%	\label{eq:df_frame_extraction_eq}
%\end{equation}
%%
%Here, $f_{\text{min}}$ is chosen as the desired number of frames we want to obtain from short and potentially dynamic videos.
%We observed that $f_{\text{min}}=600$ shows good performance.
%For videos with longer duration, the number of frames extracted increases by 10 frames per minute relatively slow.
%Thereby, we limit the storage requirements for long videos.
%For videos longer than 12 hours, we do not further increase the number of extracted frames by limiting it to $f_{\text{max}}$=8,000.
%This results in a constant increase in the number of frames extracted for videos up to about 12 hours.
%For videos with a longer duration, the number of frames extracted is limited to 8,000.
%
One might argue that the frame extraction approach used is quite simple and could benefit, e.g., from shot boundary detection or face tracking.
Therefore, we additionally tested a sophisticated approach, that does not analyze individual frames but face tracks, i.e., the face is detected once and followed through the subsequent frames.
Therefore, we used Pyannote-Video\footnote{\url{https://github.com/pyannote/pyannote-video} [Accessed: \today]} as a framework to consider a face tracking approach as well in our design.
This framework is based on the face recognition framework Openface~\cite{Amos2016} and implements shot boundary detection, face track extraction, and clustering via hierarchical agglomerative clustering.

\subsection{YouTuber Identification}\label{sec:face-clustering}
In the previous section, we discussed how a set of face images per video is obtained using Facenet.
In a following step, we use clustering to create an association between images and people.
In the case of face tracks, these are already grouped per individual. However, due to shot boundaries, i.e., scene or camera switches, multiple face tracks per individual may exist.

In a next step, we evaluate five different clustering approaches: Pyannote-video in two configurations, DBSCAN, HDBSCAN, and agglomerative clustering (AGG).
Pyannote-video can be configured with different frame extraction and face detection rates.
Therefore, two settings are chosen:
\emph{C1}: Every frame is used for face tracking and 1 frame per second is used for face recognition.
\emph{C2}: Only 1 frame per second is extracted as well as used for face recognition.
To assess the performance, we evaluate metrics for error rate, number of not clustered images, number of clusters found, and the clustering runtime.

\begin{figure}[t]
	\centering
	\includegraphics[width=0.95\linewidth]{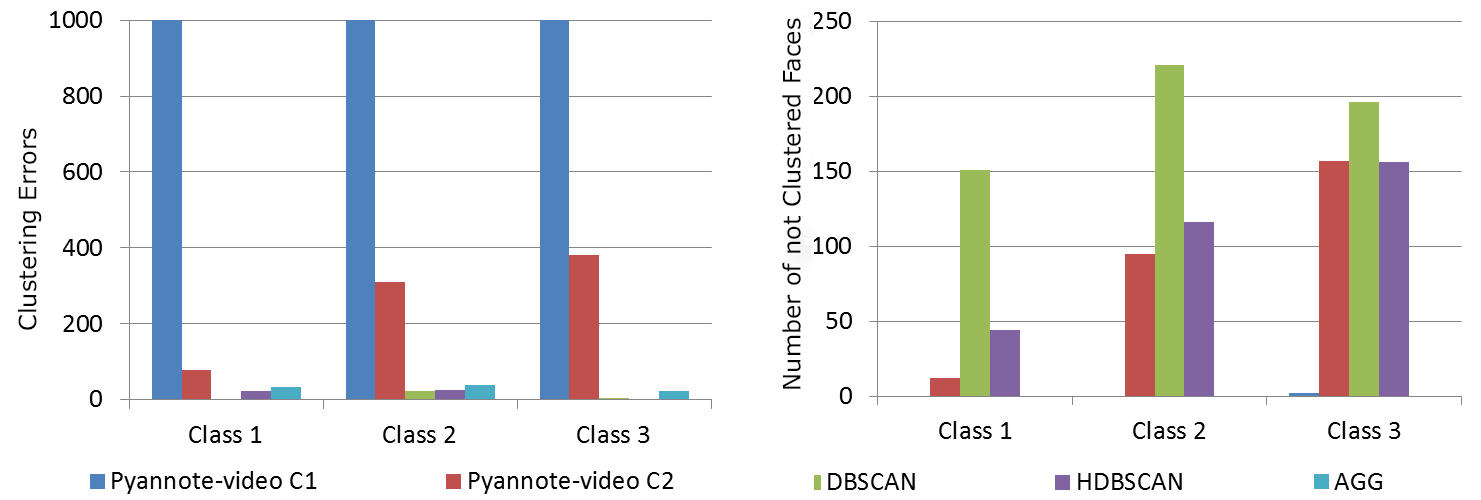}
	\caption{(Left) Number of clustering errors. (Right) Number of not clustered face images.}
	\label{fig:clustering_eval-a}
\end{figure}

\begin{figure}[t]
	\centering
	\includegraphics[width=0.95\linewidth]{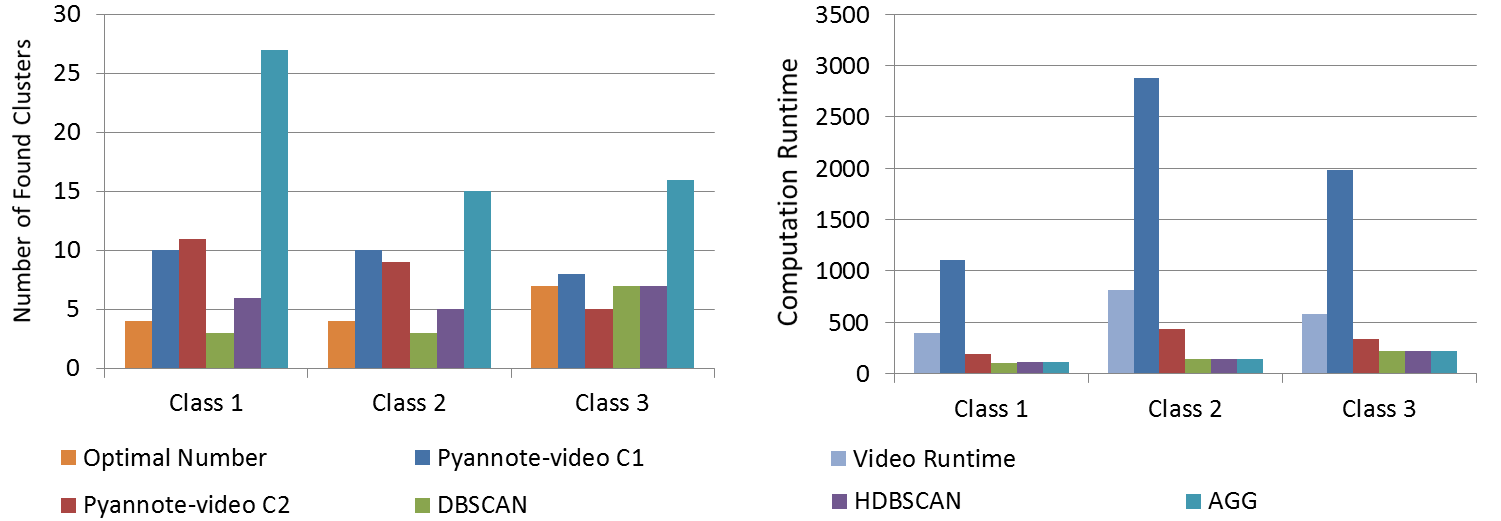}
	\caption{(Left) Number of clusters found compared to reference (optimal). (Right) clustering run-time comparison.}
	\label{fig:clustering_eval-b}
\end{figure}

\Cref{fig:clustering_eval-a} depicts the evaluation results of the approaches based on the clustering error and number of not clustered face images.
In \Cref{fig:clustering_eval-b}, we show the evaluation results based on the number of identified clusters and the clustering runtime.
We can see that Pyannote-Video C1 has a high number of erroneously clustered face images.
Furthermore, clustering run-time is multiple times higher for C1
than the video duration, and also multiple times higher than required by the other approaches.
The second configuration C2 performs better than C1, but still worse than the other approaches regarding clustering errors and clustering runtime.
Comparing the two similar approaches DBSCAN and HDBSCAN a difference in noise handling, measured by the number of not clustered face images can be noticed.
DBSCAN identifies more images as noise than HDBSCAN and all other approaches.
The number of found clusters fits for both density-based approaches: DBSCAN and HDBSCAN, being close to the optimal number.
The major advantage of both density-based approaches, i.e., DBSCAN and HDBSCAN over the other approaches is that no estimation of the number of cluster must be given.
Finally, we conclude that both DBSCAN and HDBSCAN perform best amongst the evaluated techniques.
The face tracking approach, while promising in theory, did not perform well.
As the number of not clustered face images suggests, DBSCAN handles noise more conservative than HDBSCAN, resulting in possible useful images being discarded.
A significant advantage of HDBSCAN over DBSCAN, besides handling varying cluster density, is that it returns probabilities describing the strength of membership for every face image.
This information can be leveraged to further filter outliers and improve the clustering results.
\emph{For these reasons we decided to use HDBSCAN for clustering the face images.}

While evaluating the performance measures above, we noticed non-compliant behavior of HDBSCAN for class one videos, especially in which no other (external) actor appeared.
HDBSCAN thereby fails to create a single cluster and labels all data as noise. This observed effect is due to the hierarchical approach and it could not be improved through parameter modification.
As DBSCAN performs well on a single cluster case, we use DBSCAN as a fallback solution, coming into effect when HDBSCAN fails to detect a single cluster.
%
%% SHORTEN and SUMMARIZE THE FOLLOWING IF SPACE NEEDED
%

An alternative approach  to clustering is classification.
However, this approach was discarded, as it does not fit our requirement that the face recognition technique should be able to distinguish between individuals without prior training data.
For the sake of completeness, we will give a brief overview of this alternative approach.
Through classification, it is possible to assign detected faces to known individuals.
The problem with this approach in our setting is that training data is needed for classification.
As we have channel information, which is analyzed beforehand, we could acquire training images through, for example, Google image search.
The drawback with this approach is that the number of content creators per channel is unknown.
Thus, the automatic retrieval of correct training data for multiple persons through image search is not viable.
Furthermore, appearing individuals that are not related to any analyzed channel would not be recognized.
For these reasons, the previously described clustering approach is utilized, incorporating both, HDBSCAN and DBSCAN.
%Thereby, clustering can differentiate between individuals without any training data.

\subsection{Collaboration Detection}\label{collaboration_detection}
We define a collaboration as the co-occurrence of a YouTuber from a different channel in a YouTuber's video, e.g., in a video showing both YouTubers or playing a (potentially) prerecorded clip of the featured YouTuber.
To identify collaborations, we build connections between videos using the previously derived face clusters.
Single videos may have multiple face clusters, i.e., at least one per detected individual.
Therefore, we compute a similarity matrix between face clusters using the Euclidean distance as a similarity measure.
In a next step, the similarity matrix is taken as an input for HDBSCAN clustering which groups all face clusters of individual persons based on the similarity measure.
This gives us face cluster-wise connections showing that the same person appeared in all connected videos.
%At this point, we know which person appeared in which video and, hence, on which channel.
This  information allows the investigation of collaborations between different channels.

Although we can now connect the appearing individuals, we do not know yet which of the persons is a content creator, i.e., owner of the channel, or an external actor appearing.
To determine the content creator of a channel, we can select the face cluster with the highest number of appearances on the channel.
However, this approach would be unable to assign multiple content creators to a single channel.
As do not want to restrict this method to single content creator channels and we assume that some
% leider keine Referenz gefunden :(
%that a large
portion of the available channels already have multiple content creator, we use the following approach.
To decide if an individual is a content creator, we leverage the number of appearances per individual and channel.
In detail, a person may have appeared in different channels, we assign the person as a content creator for the channel with the highest number of appearances.
Thereby, a channel with one or multiple content creators can be correctly detected.
%TODO Christian: Evtl. Satz zur manuellen Verifikation des Ansates (200 Channels von Hand ??berpr??ft)
\begin{figure}[h]
	\centering
	\includegraphics[width=0.6\linewidth]{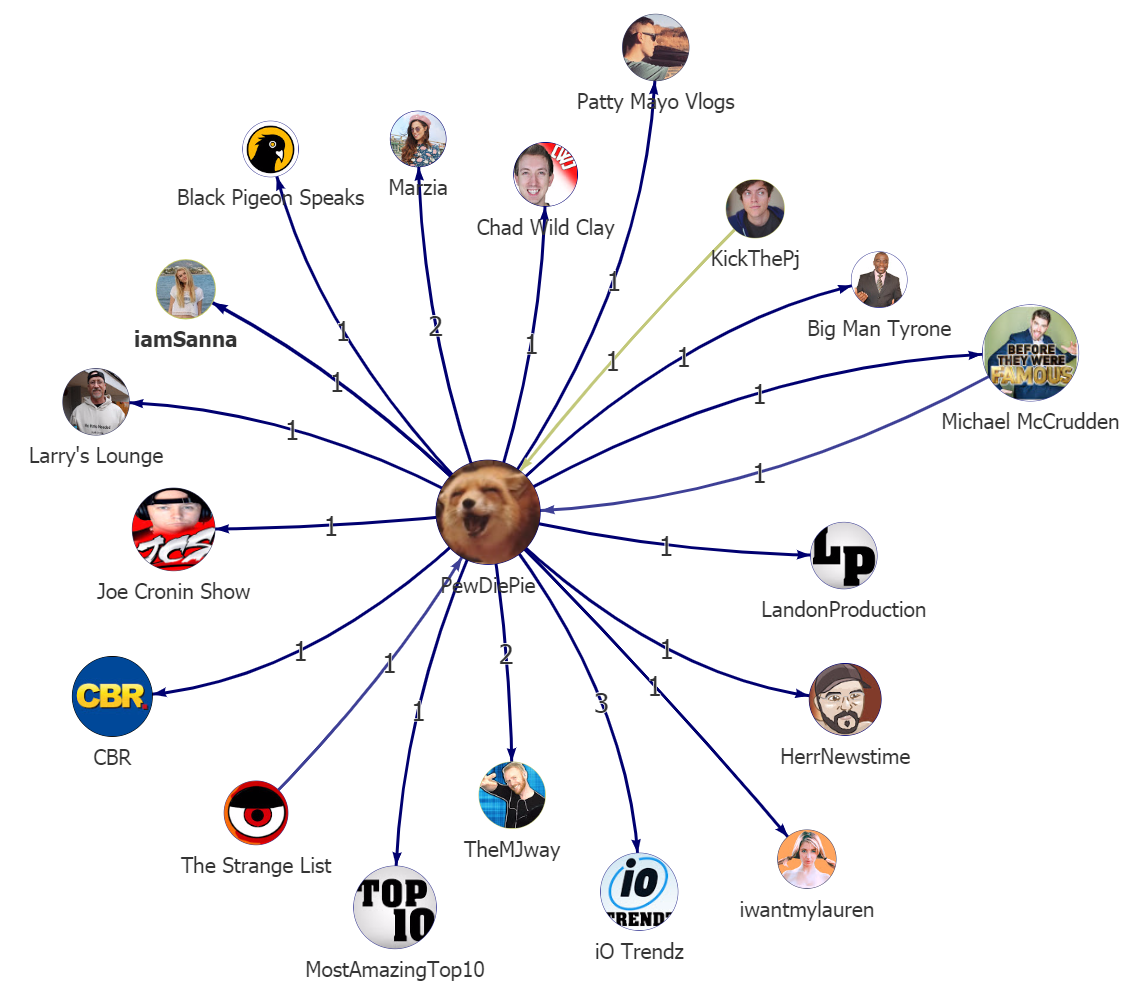}
	\caption{Sample graph of PewDiePie's YouTube channel and 1-hop neighbors\protect\footnotemark }
	\label{fig:sample-collaboration-graph}
\end{figure}
\footnotetext{Channel images taken from https://www.youtube.com/\{PewDiePie,
chadwildclay,
channel/UCEYLdM2bdhmw-TS3c0TjFNw,
KickThePj,
TheBigManTyrone,
MrMccruddenmichael,
user/LandonProduction,
HerrNewstime,
iwantmylauren,
iOTrendz,
channel/UCIq3bpW-MaAzj4Y2G9ezPhA,
channel/UCBINYCmwE29fBXCpUI8DgTA,
channel/UCm3GpkVRonpt2BHrt3GhwjQ,
comicbookresources,
JoeCroninSHOW,
channel/UCZApDB9BW7ZjNcPb3Wu7rRg,
imsannachanel,
TokyoAtomic,
CutiePieMarzia\}}
\begin{comment}
\footnotetext{YouTube channel images taken from: https://www.youtube.com/PewDiePie, https://www.youtube.com/chadwildclay, Patty Mayo: https://www.youtube.com/channel/UCEYLdM2bdhmw-TS3c0TjFNw, https://www.youtube.com/KickThePj, https://www.youtube.com/TheBigManTyrone, https://www.youtube.com/MrMccruddenmichael, https://www.youtube.com/user/LandonProduction,
https://www.youtube.com/HerrNewstime,
https://www.youtube.com/iwantmylauren,
https://www.youtube.com/iOTrendz,
https://www.youtube.com/channel/UCIq3bpW-MaAzj4Y2G9ezPhA,
https://www.youtube.com/channel/UCBINYCmwE29fBXCpUI8DgTA,
https://www.youtube.com/channel/UCm3GpkVRonpt2BHrt3GhwjQ,
https://www.youtube.com/comicbookresources,
https://www.youtube.com/JoeCroninSHOW.
https://www.youtube.com/channel/UCZApDB9BW7ZjNcPb3Wu7rRg,
https://www.youtube.com/imsannachanel,
https://www.youtube.com/TokyoAtomic,
https://www.youtube.com/CutiePieMarzia
}
\end{comment}
\subsubsection*{Collaboration Graph}
Inspired by \cite{Gugel2015, Wilson2009}, the collaborations between channels are modeled as a graph with channels as nodes and collaborations as directed edges connecting the nodes.
%Thereby, collaborations are modeled as directed edges.
The edge direction describes that a content creator of the origin channel appeared, i.e., collaborated, in one or more videos of the destination channel.
\Cref{fig:sample-collaboration-graph} depicts an example graph showing one the world's most popular YouTubers: PewDiePie.
%Each neighboring node denotes a YouTube channel in which PewDiePie appeared if the directed edge is pointing towards the other channel.
%If the directed edge points towards PewDiePie, the other YouTuber appeared in one of his videos.
%We observe just one channel for which both is the case, i.e., Michael McCrudden.
The edge label denotes how often collaborations between the two channels were observed.
We use a graph to visualize and model the collaborations as we can apply different graph algorithms to analyze the underlying channel relations.
%Furthermore, it is possible to visualize the derived collaborations in a graph.
%TODO Hier dann Verweis auf das online tool

\begin{table}[t]
\centering
    \footnotesize
	\caption{Overview of the available crawler tools, $^{\ast}$are outdated\label{tab:bg_data_crawler}}
	\begin{tabular}{ l  l  l l  l  l l }
		\toprule
		\textbf{} & \textbf{Scrapy}        & \textbf{YTCrawl} & \textbf{YOUStatAnalyzer} & \textbf{HarVis} & \textbf{ytdata$^{\ast}$} & \textbf{TubeKit$^{\ast}$}\\ \midrule
		\textbf{Customizable}  & Highly           & limited                  & limited         & limited         & limited          & limited  \\
		\textbf{Documentation} & Extensive        & sparse                   & sparse          & sparse          & sparse           & sparse  \\
		\textbf{Storage}       & Customizable     & File                     & MongoDB         & SQL             & SQLite           & MySQL    \\
		\textbf{API usage}     & Yes              & No                       & Yes             & Yes             & Yes              & Yes      \\
		\textbf{Language}      & Python           & Python                   & Python          & Java            & Python           & PHP      \\ \bottomrule
	\end{tabular}
	
\end{table}

%=============================================
\section{YouTube Statistics Data Acquisition}\label{data_aquisition}
%=============================================
In the following, we briefly describe how we crawl the required information from YouTube and specify how we select appropriate YouTube channels for crawling our dataset.
%As the goal of this work is on identifying and analyzing collaborations, random crawling of YouTube channels is not an appropriate method for data acquisition.
%Instead, channels that are likely to collaborate are desired.

\subsection{YouTube Data Crawling}
\label{sec:im_youtube_scraper}

To assess the effect of YouTuber collaborations, we need to acquire channel metadata such as the subscriber count and video view counts.
Additionally, we also want to acquire the video and channel metadata, such as video titles, video and channel descriptions, and the channels' featured channel list.
%We design a data crawler.
%To this end, an analysis of existing crawling tools is conducted and the respective candidates introduced in the following.
%%
%\textbf{YTCrawl}\fnurl{https://github.com/yuhonglin/YTCrawl} is a YouTube video history data crawler, used in~\cite{Yu2014}.
%Its purpose is to extract the video view count history as well as histories for the subscriber counts, share counts, and watch-time of videos.
%YTCrawl supports single and batch crawling methods with a crawling rate of 36,000 videos/hour.
%Whereas the history data of videos is not available for all videos, publishing this data is up to the uploader and~\cite{YTCrawlGH} reports that only about 60\% of videos have this history accessible.
%%
%\textbf{YOUStatAnalyzer}\fnurl{https://github.com/mattiazeni/youstatanalyzer} is a crawling tool for extracting popularity metrics from YouTube videos, for example, view count, like, and dislike count as well as metadata, e.g., video title.
%Proposed in~\cite{Zeni2013}, its purpose is to create a video-statistics database for video analysis.
%%
%\textbf{HarVis} is a complete framework for YouTube statistics analysis, including data crawling, storage, and visualization.
%Demonstrated in \cite{Ahmad2016} by analyzing data for the in this time popular "Gangnam Style"-themed videos.
%%
%
The crawled data is not filtered by a specific video topic but through a seed set of YouTube channels and their uploaded videos.
Under those requirements most of the existing tools such as \textbf{YTCrawl}\footnote{\url{https://github.com/yuhonglin/YTCrawl} [Accessed: \today]}, \textbf{YOUStatAnalyzer}\footnote{\url{https://github.com/mattiazeni/youstatanalyzer} [Accessed: \today]}, or \textbf{HarVis}\footnote{\url{https://github.com/DrUzair/HarVis} [Accessed: \today]} cannot be leveraged for our purpose without significant code changes.
Customizing the frameworks requires a solid documentation, which most of the tools do not provide or only in sparse form.
To this end, we decide to develop a YouTube data crawling tool based on the \textbf{Scrapy}\footnote{\url{https://scrapy.org} [Accessed: \today]} framework which is highly customizable and provides extensive documentation, that allows to implement all requirements of our envisioned crawler in reasonable time and with a comparable small effort.
Scrapy is widely used and has a big open source community.
\Cref{tab:bg_data_crawler} shows a qualitative comparison of the considered YouTube data crawling tools, including Scrappy.

\begin{figure}[h]
	\centering
	\includegraphics[width=0.95\textwidth]{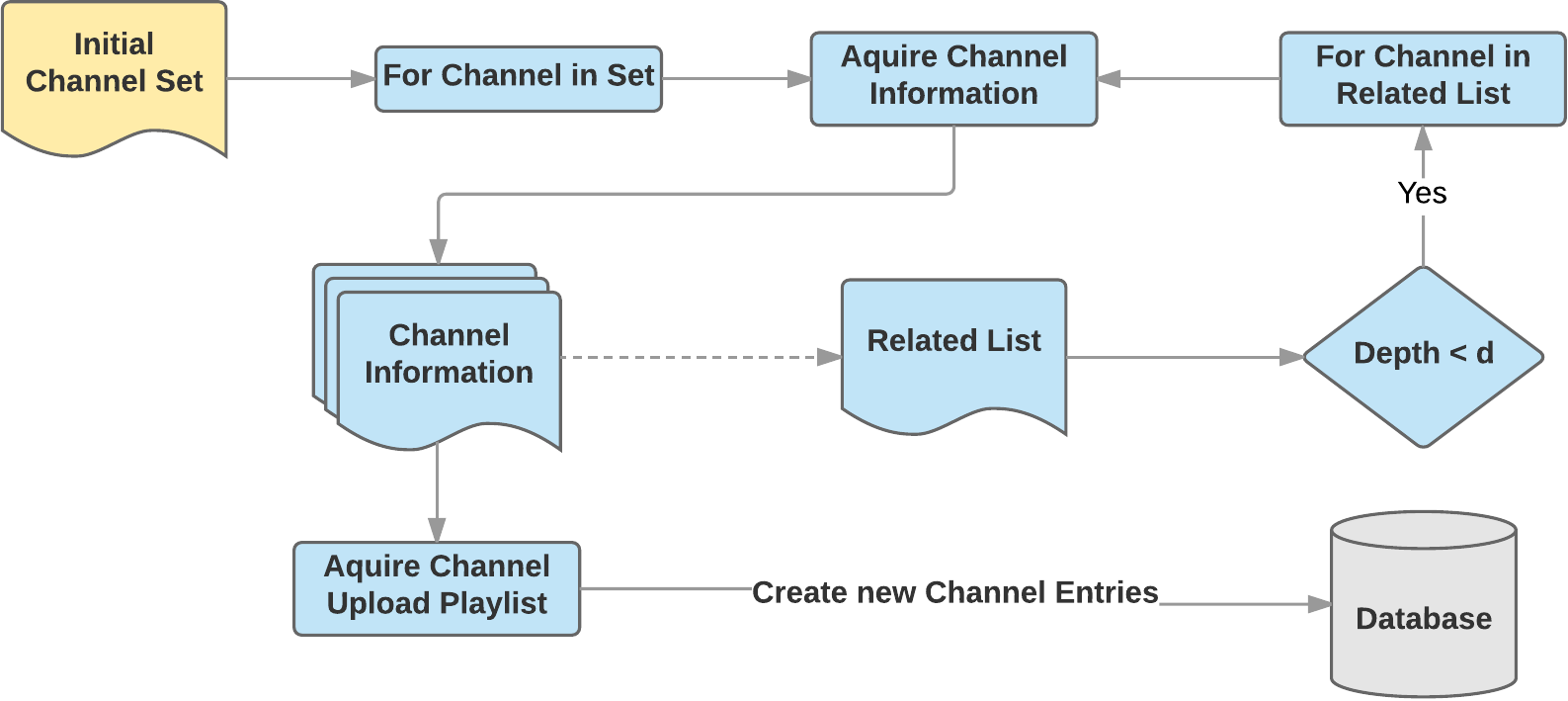}
	\caption{Populate spider architecture, d is the user-defined maximum depth for channel recursion.}
	\label{fig:im_populate_spider}
\end{figure}

Next, we describe the design of the Scrapy-based crawling architecture.
As we distinguish between continuous and static data, we implemented two spiders, i.e., crawling modules in Scrapy.
The first spider, which we denote as \emph{Populate Spider}, crawls static data and populates a database with channel and video entries, e.g., the video and the channel name.
\Cref{fig:im_populate_spider} illustrates the crawling process.
The populate spider takes an initial set of channel IDs for which it requests all static data and subsequently creates a database entry for every crawled channel.
Note that for fixed set of channels $\bm{d}$ is set to zero to hinder Scrapy from adding additional channels to the set.

In the next step, a second spider denoted \emph{Daily Spider} crawls continuous data created by the channels and videos already known in a daily manner, i.e., view count and subscriber count.
Here, each channel's \emph{Upload Playlist} is crawled, which contains the videos uploaded by a channel to identify newly uploaded videos.
Crawling the popularity statistics for all videos uploaded on the given channels allows constructing time series of popularity statistics.
We record all interactions on the monitored channels such as view,  comment, and subscriber counts in the time span between 28.12.2016 and 28.03.2017 on a daily basis.

\begin{figure}[t]
	\centering
	\includegraphics[width=0.95\textwidth]{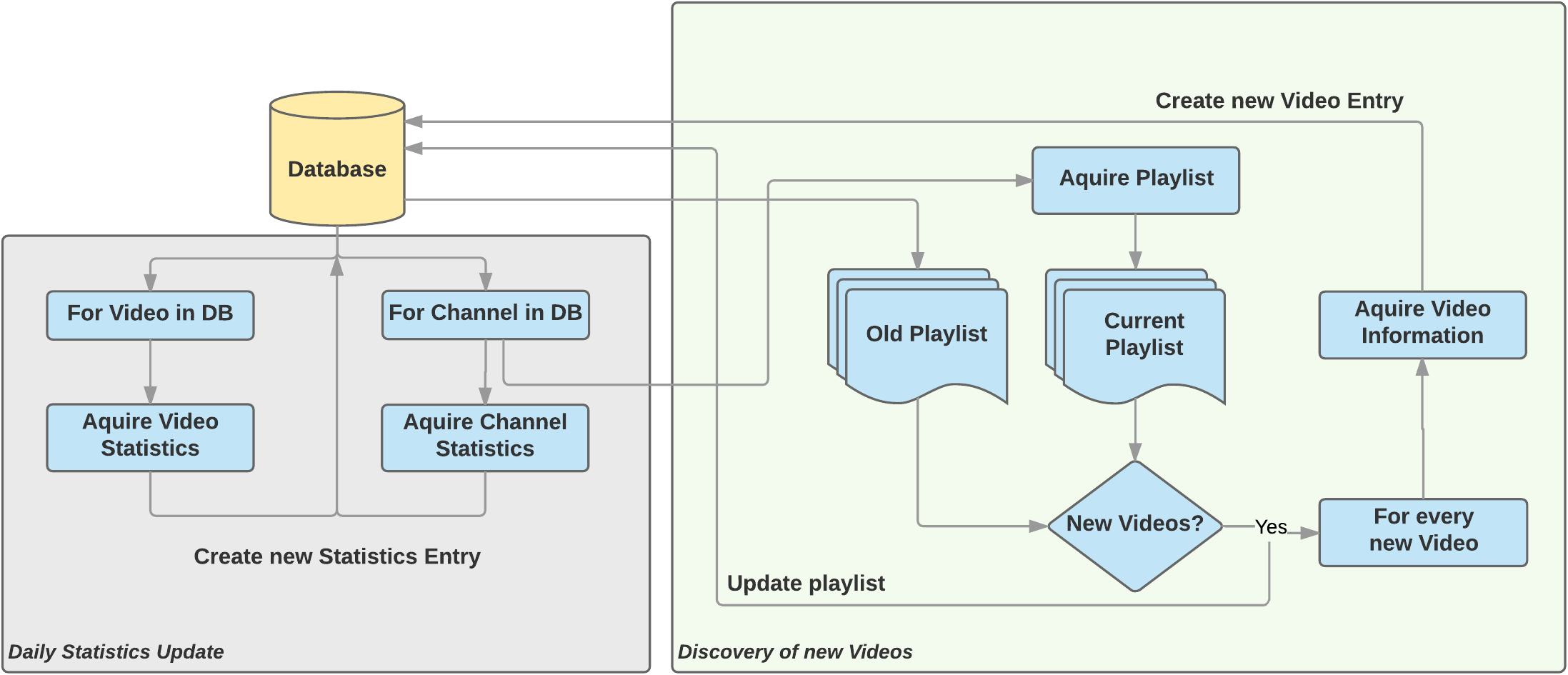}
	\caption{Daily spider architecture.}
	\label{fig:im_daily_spider}
\end{figure}

\subsection{Data Selection and Representation}
\label{subsec:data_selection}

% Channel Selection
The \emph{Populate Spider} needs an initial set of YouTube channel IDs.
To create a qualitative and large dataset of channels, we crawl three of the most popular MCNs~\cite{SocialBlade} as channels associated to a MCN have higher chances to collaborate~\cite{TubeFilter2015}.
These MCNs are described in \cref{tab:bg_networks_worldwide}.
We used the website SocialBlade\footnote{\url{https://socialblade.com/youtube/top/networks/most-subscribed} [Accessed: \today]} to derive the mapping of channels to MCN member lists.
Here, we take a random sample of $1.5\times10^3$ channels for every MCN in which the 100 channels with most subscribers per MCN are included.
%
% Features Channel List
In a next step, we use the so called \emph{Featured Channel List} of the crawled channels, which contains other channels defined by the channel owner to express an acquaintanceship between his and other channels.
This relationship is modeled in a graph representation as described in \cref{collaboration_detection}, containing roughly 44k channels.
Here, channels are represented by nodes, while the featured channel list comprises unidirectional edges from one node to other nodes.
\begin{table}[b]
	\caption{Most popular YouTube MCNs worldwide, measured by number of subscribers.}  \label{tab:bg_networks_worldwide}
	\centering
	\footnotesize
	\begin{tabular}{ l  l  l  l  l }
		\toprule
		\textbf{Name} & \textbf{Members} & \textbf{Subscribers (over 30 days)} & \textbf{Views (over 30 days)} \\ \midrule
		BroadbandTV & 237,235 & 82,718,601 & 20,107,687,476 \\
		Studio71 & 13,194 & 18,535,751 & 5,402,265,290 \\
		Maker Studios & 9,460 & 14,011,532 & 4,613,091,2598\\ \bottomrule
	\end{tabular}
\end{table}
Finally, non-mutual edges between channels are removed, while mutual edges, i.e., nodes having each other in their \emph{Featured Channel List}, are considered likely to collaborate and, hence, are kept.
We also excluded Gaming videos as we observed a lack of face presence in these videos and, furthermore, often high resolution game figures depicted on covers or within the videos would have led to increases imprecision. 
From the remaining subgraph, we extract the largest connected component, resulting in a graph with roughly 8k nodes and about 10k edges indicating potential collaborations.
By applying CATANA to this subgraph, i.e., analyzing 2.4 years of video uploaded on these channels for a period of three months, we identified  1,599 nodes and 1,728 edges representing actual collaborations.
The edge sum, which corresponds to the overall number of collaborations sums up to 3,925, see \Cref{collaborations_table}.
\begin{table}[h]
\footnotesize
\centering
\caption{Collaborations observed in the three month's time span.}
\label{collaborations_table}
\begin{tabular}{cccccc}
\toprule
Collaborations & Duration & Mean & Median & 75-percentile & Max\\ \midrule
3,925          & 3 months & 2.8  & 1.0    & 3.0           & 134 \\ \bottomrule
\end{tabular}
\end{table}

%===============================
\section{Evaluation}\label{eval}
%===============================
In this section, we formulate and answer research questions with respect to the focus points of: \emph{(i)} collaboration frequency and partner selectivity (\Cref{subsec:collab_freq}), \emph{(ii)} the influence of multi-channel networks (MCNs) on channel collaborations (\Cref{subsec:mcn_influence}), \emph{(iii)} collaborating channel types (\Cref{subsec:channel_types}), and \emph{(iv)} the impact of collaborations on video and channel popularity (\Cref{subsec:collab_impact_on_popularity}).

%=============================================================================
\subsection{How often do collaborations happen and reoccur with the same partner?}
\label{subsec:collab_freq}
%=============================================================================
\begin{figure}[h]
	\centering
	\subfloat[Number of pair-wise collaborations]{\includegraphics[width=0.45\linewidth]{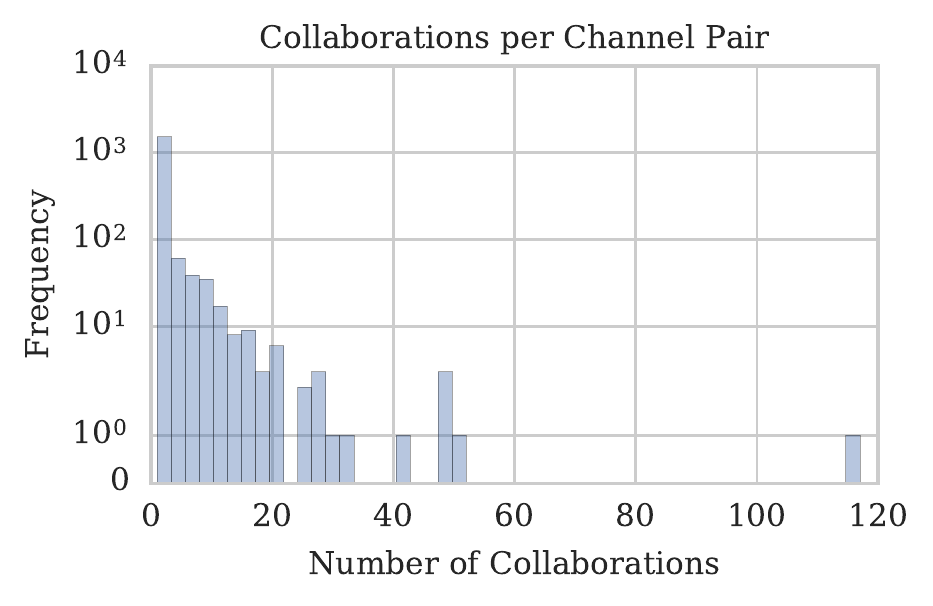}\label{fig:ev_collab_pairwise}}
	\hfil
	\subfloat[Number of collaborations per channel]{\includegraphics[width=0.45\linewidth]{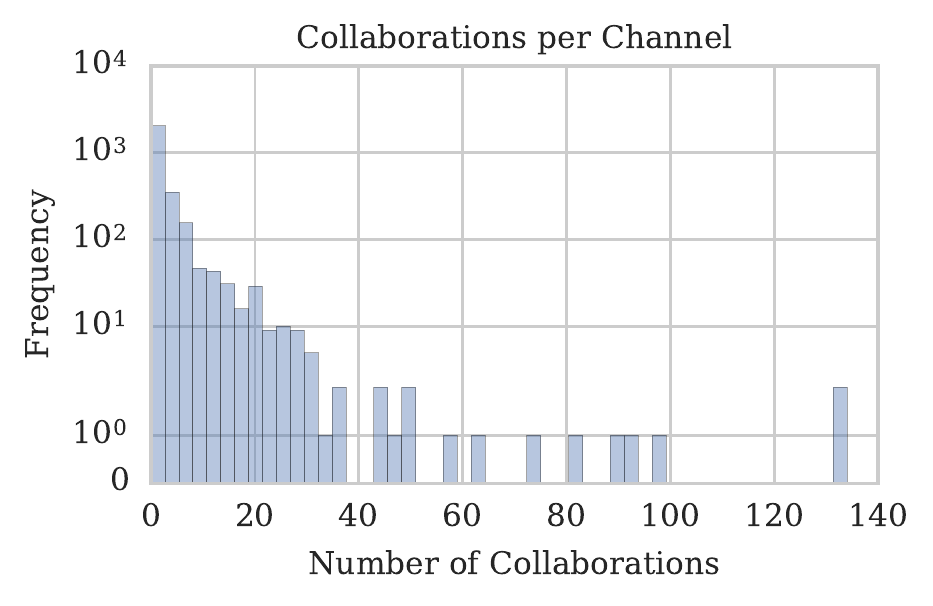}\label{fig:collab_nof_collabs_per_channel}}
	\caption{Histograms of detected collaborations per channel pairs (a) and per channel (b)\label{fig:ev_collab_numbers}}
\end{figure}

Using the collaboration graphs, we can easily determine the overall number of collaborations by summing  up the edge weights between two YouTube channels.
Analyzing the edge weights allows us to examine the number of repeated collaborations.
\Cref{collaborations_table} depicts the derived collaboration statistics where we observe that the distribution is skewed to the right, indicating the presence of a few channel pairs with a comparably high number of collaborations.
%Here, we can see that 50\% of identified collaborations between channels only occurred once, while 75\% occurred at most twice.
%As the mean is higher than the median, the distribution is skewed to the right, indicating the presence of a few channel pairs with a high number of collaborations.
We confirm this by drawing the histogram of the collaboration counts in \Cref{fig:ev_collab_pairwise}, showing the distribution of repeated collaborations between two YouTubers who have collaborated at least one.
% which indicates a power-law-distribution.
We deduce that over a 3-months observation period collaborations between two channels rarely happen more than once.

In a next step, we analyze the number of collaborations per channel instead of distinct channel pairs.
Therefore, we sum up the edge weights for every channel node.
Taking the perspective of a channel, we differentiate between collaborations taking place in own videos (internal) and videos of other channels (external).
If a collaboration between YouTuber A and B occurs on YouTuber A's channel, it is considered as an internal collaboration by A and as an external collaboration by B.
This distinction helps us later to detail the effects on both sides of a collaboration.
\Cref{fig:collab_nof_collabs_per_channel} shows the corresponding distributions.
%We note that the median of the outgoing number of collaborations is \textbf{zero}, whereas the median of the ingoing ones is \textbf{one}.
%The mean for both in- and outgoing is, however \textbf{1.74}, as for every outgoing value, an ingoing one exists.
%The median indicates that there are central channels, with a high number of outgoing collaborations directed at multiple channels with only a single ingoing collaboration.
We conclude that a small number of highly collaborating channels denoted \emph{central channels} exists, with most of the remaining channels having very few collaborations.
%having collaborations with a large number of other channels that collaborate only with one of the highly collaborative channels.
%These channels have a high number of collaboration links with channels that exhibit only a single collaboration with them self.
Thus, central channels show a high in-degree, demonstrating a key \emph{influencer} role on YouTube.
%Such central channels are likely to be highly influential as they appear on multiple other YouTubers' channels and, therefore, are known to a large audience base.
These YouTubers are especially valuable for product placements, advertisements, and, hence, are especially valuable assets for their MCNs.
\begin{figure}[b]
	\centering
	\subfloat[Collaborations per uploaded videos ]{\includegraphics[width=0.45\linewidth]{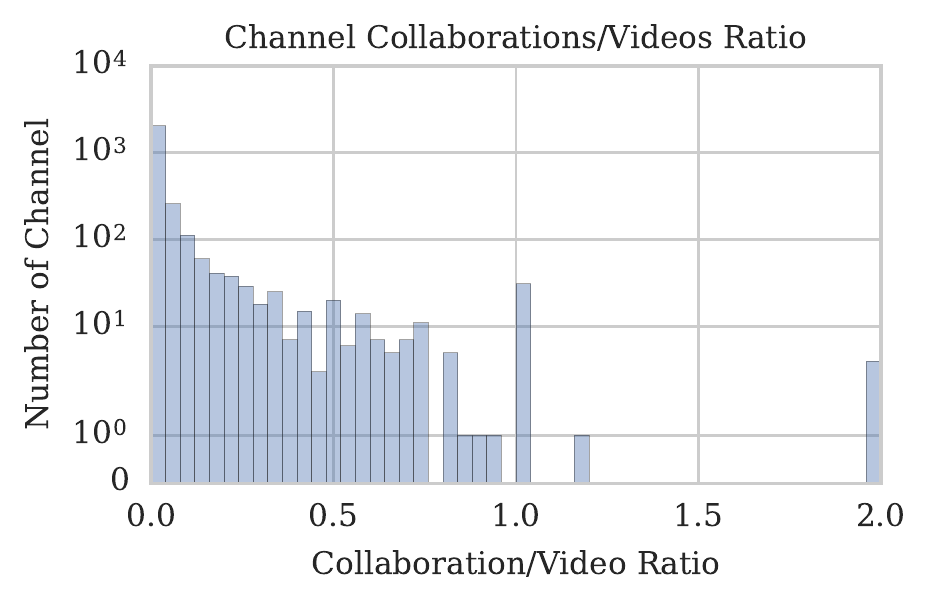}\label{fig:collab_ratio}}
	\hfil
	\subfloat[Channel collaborations]{\includegraphics[width=0.4\linewidth]{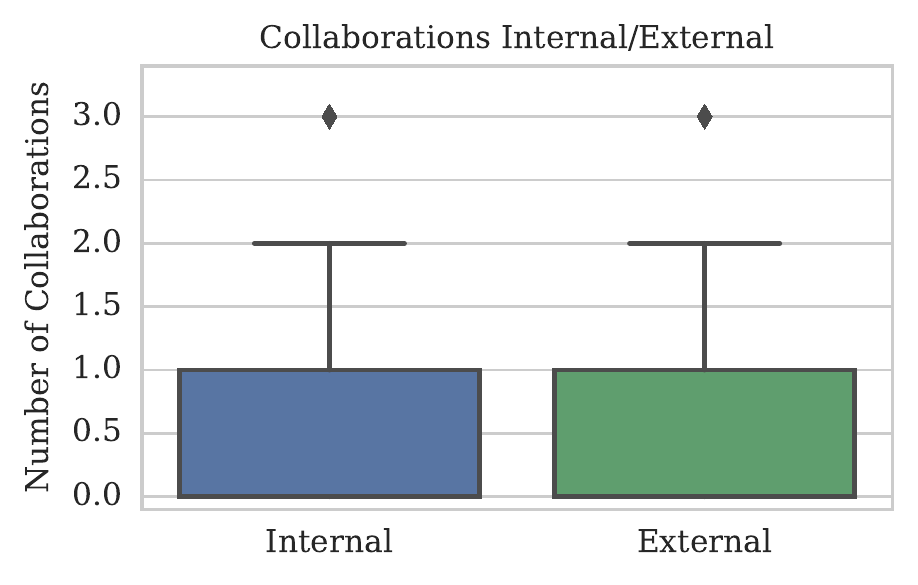}\label{fig:collab_in_out_collabs}}
	\caption{Collaboration/Video ratio and separated in- and outgoing collaborations\label{fig:ev_collab_ratio_in_out}. }
\end{figure}
One weakness of the former analysis is that it compares results based on the absolute number of collaborations.
%This leads to a positive bias for channels with more uploads but lacks on relative comparability.
Thus, we investigate the relative ratio of a YouTuber's collaborations compared to the overall number of the channel uploads.
Given the number of internal collaborations of a channel $k$, denoting collaborations only occurring in its own videos, and the number of videos of the channel $n$, we calculate the collaboration ratio as $k/n$.
\Cref{fig:collab_ratio} shows the distribution of the collaboration ratio.
Values around one imply that in nearly every video of the channel, a collaboration is found.
This may indicate that certain content creators regularly work together, or share a common channel while also operating separate channels alone.
Values larger than one can occur if multiple collaborations were detected in a single video.
%, for example because multiple content creators are present.
%\textbf{@Moritz: Fig 8.b passt nicht mehr zum Text. Warum? @Christian: Wenn man sich die verteilung in zahlen ansieht haben wir immernoch ein 50\% von 0.0 und ein 75\% quantil mit 1.0, es liegen also immernoch die meisten nahe bei null/kleiner eins, kommt im boxplot glaube besser zu Geltung da es dort viel enger um 0.0 und 1.0 dargestellt wird:}
In \Cref{fig:collab_in_out_collabs}, we observe only a few outliers with a ratio above one, while most of the channels have a ratio close to zero.
This indicates that a large portion of channels have only a single ingoing collaboration.
%We confirm this by the high number of channels having a collaboration count near zero as depicted in \Cref{fig:collab_ratio}.
Summarizing our finding with respect to collaboration frequency we observe that a single channel collaborates with other channels on average 2.8 times ($[2.5-3.15]$ at a 99\% confidence level).
%As our data covers three months, this indicates on average 1.17 collaborations per month and channel.
If channels collaborate, we find in our 3-month dataset that they repeat their collaboration on average 2.3 times ($[2.0-2.6]$ at a 99\% confidence level).
%in our dataset, while the 99\% confidence interval ranges between 1.87 and 2.33.
%This results in 0.7 collaborations per channel pair and month on average.
Overall, the distributions for collaboration metrics are skewed as a consequence of a few highly influential YouTubers.

%Central channels thereby describe a similar situation as the sample group displayed in Fig.~\ref{fig:ev_collab_sample}.
%\begin{figure}[!tb]
	%\centering
	%\subfloat[Mean of pair-wise collaborations.]{\includegraphics[width=0.4\linewidth]%{figures/collab_channel_pairs_bar_box.pdf}\label{fig:ev_collab_pairwise_bar}}
	%\hfil
	%\subfloat[Mean of collaborations per channel.]{\includegraphics[width=0.4\linewidth]{figures/collab_nof_collabs_per_channel_bar_box.pdf}\label{fig:collab_nof_collabs_per_channel_bar}}
	%\caption{99\% confidence intervals for number of collaborations\label{fig:ev_collab_pairwise_nof}}
%\end{figure}

%=============================================================================
\subsection{How do multi-channel networks (MCNs) influence channel collaborations?}
\label{subsec:mcn_influence}
%=============================================================================

\begin{figure}[h]
	\centering
	\includegraphics[width=.9\linewidth]{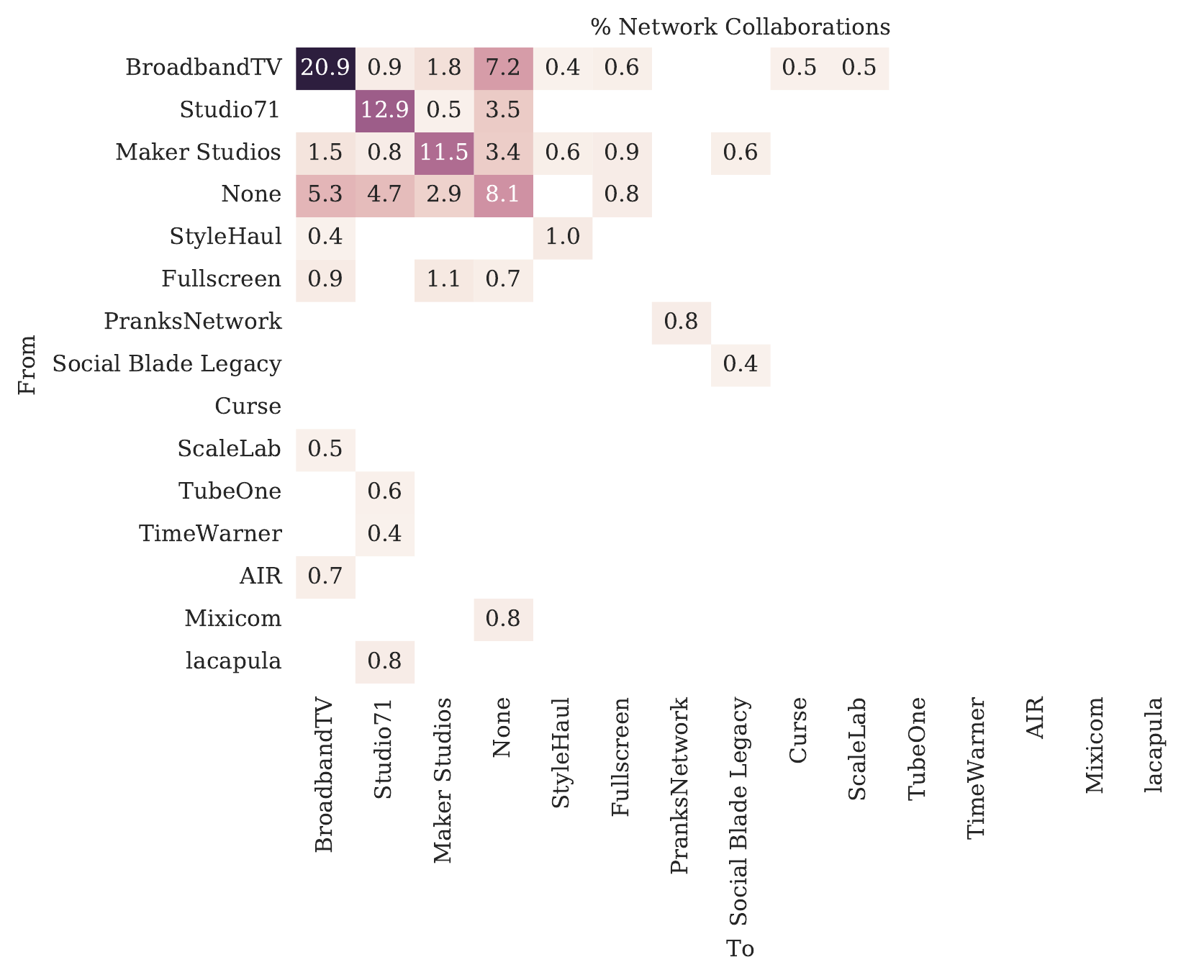}
	\caption{Absolute number of collaborations within and between MCNs.}
	\label{fig:ev_collab_network_map}
\end{figure}

To answer this question, we first analyze which collaborations take place between MCNs.
To this end, we first determine the respective MCN for each channel using publicly available information\footnote{\url{https://socialblade.com/} [Accessed: \today]} to augment the channel information of our collaboration graph.
\Cref{fig:ev_collab_network_map} depicts the most collaborating MCNs in form of a MCN-collaboration matrix.
Overall, we found 405 MCN pairs collaborating with each other.
% -- most of them only once or twice.
Note that the entry \emph{None} refers to channels for which we could not determine a MCN association.
%The figure only includes networks with more than 10 collaborations for a better overview and clarity.

\begin{figure}[t]
	\centering
	\includegraphics[width=0.85\linewidth]{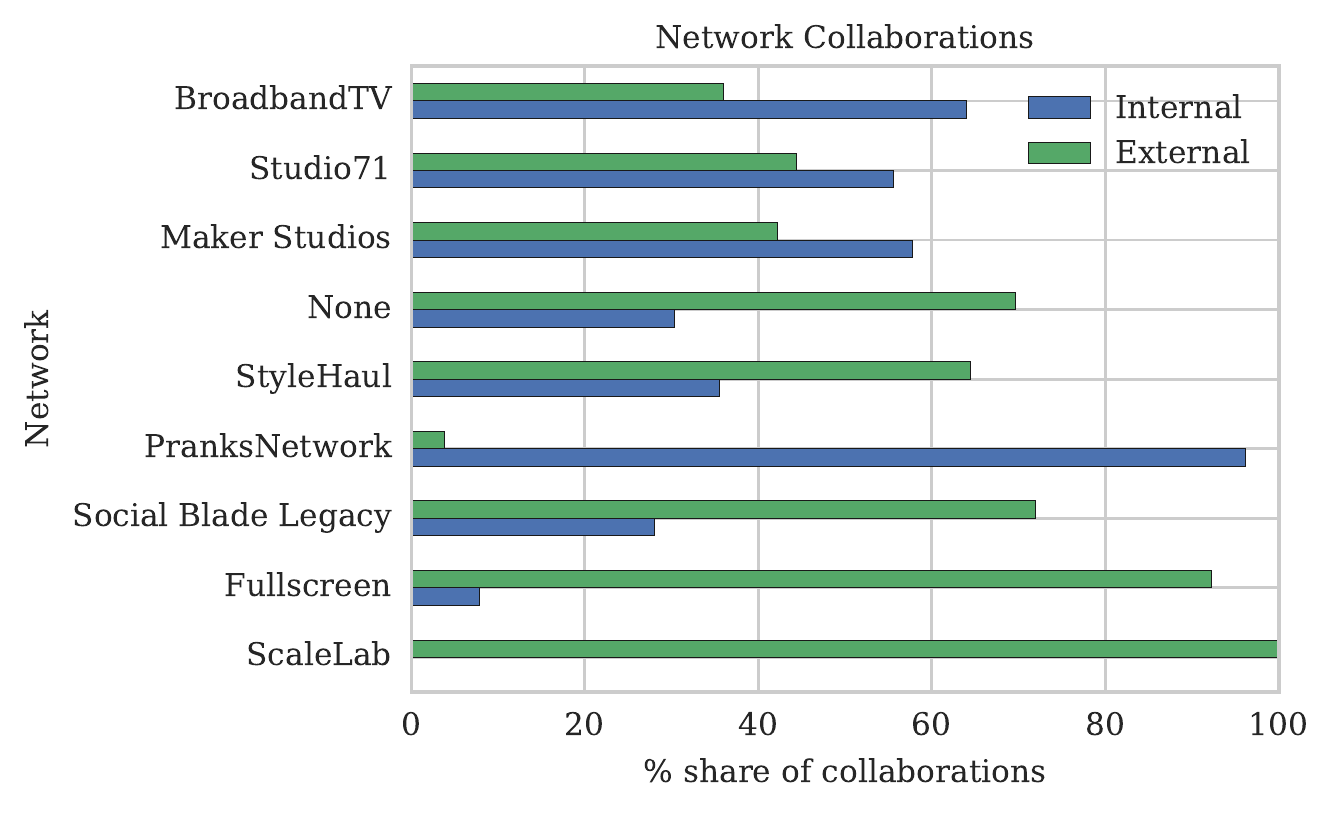}
	\caption{Internal and external MCN collaborations.}
	\label{fig:ev_collab_network_in_out}
\end{figure}

Examining the diagonal of the matrix in \Cref{fig:ev_collab_network_map} we observe a distinct trend showing that most collaborations occur within MCNs, and thus between their members.
%We An exception is the entry \emph{None}, which refers to channels without an associae ted MCN.
Further, we observe that significant collaborations between networks are mainly confined to the three dominant networks, namely, BroadbandTV, Studio71, and Maker Studios.
We find many collaborations of unassociated channels, i.e., with the label \emph{None}, with the three dominant networks, which we attribute to the fact that they are the world's three largest MCNs and, hence, have YouTube channels associated which are popular and an attractive target for collaboration.
Furthermore, famous YouTubers are a popular topic for other YouTuber's that may show  the popular YouTuber's face or video sequences.
%
%Most collaborations occur between the three dominant networks, namely, Studio71, BroadbandTV, Maker Studios, and the unassociated \emph{None}.
%This is plausible as they cover the largest portion of channels, see \Cref{tab:bg_networks_worldwide}.
%
\begin{comment}
as seen in Fig.~\ref{fig:ev_network_member}.
\begin{figure}[!tb]
\centering
\includegraphics[width=0.4\textwidth]{figures/network_nof_channel.pdf}
\caption{Network channel member.}
\label{fig:ev_network_member}
\end{figure}
\end{comment}
%
%Please note that besides the collaborations within the largest MCNs and \emph{None}, no significant interaction between different networks could be observed.
We deduce that belonging to a MCN strongly increases the probability of a YouTuber to collaborate.

\Cref{fig:ev_collab_network_in_out} shows the percentage share of collaborations separated by MCN in- and external collaborations.
Here, we observe that the three largest MCNs, i.e., BroadbandTV, Studio71, and Maker Studios as well as PranksNetwork have much more internal collaborations compared with the smaller MCNs.
Hence, their YouTubers collaborate more in their own videos than on the videos of other MCN's YouTubers.
Note that a large portion of the outgoing collaborations is with channels that are not associated with a MCN.
Channels not belonging to a MCN show a preference to work with MCN-associated channels as more than 70 percent of their collaborations are outgoing.
We conclude that an influence of MCNs concerning YouTuber collaborations can be inferred.
In summary, we find that channels associated with a MCN collaborate more often with each other and if collaborations occur outside the MCN, then they are usually with non-associated channels and rarely with other MCNs' YouTubers.
%Is is not far to seek the reason of this behavior in MCNs discouraging collaborations with other potentially competing MCNs.

\begin{comment}
However, a further factor should be considered, as we retrieved the network association data from an external source, namely SocialBlade, it is possible that the data is outdated.
Consequently, a portion of un-associated could presently be associated with a network, or associated channel may have changed networks. When considering this, detected collaborations between different networks or un-associated channels may occurred while they were actually within the same network.
\end{comment}

%=============================================================================
\subsection{Which channel types collaborate?}
\label{subsec:channel_types}
%=============================================================================

\begin{figure}[h]
	\centering
	\subfloat[Popularity classes.]{\includegraphics[width=0.35\linewidth]{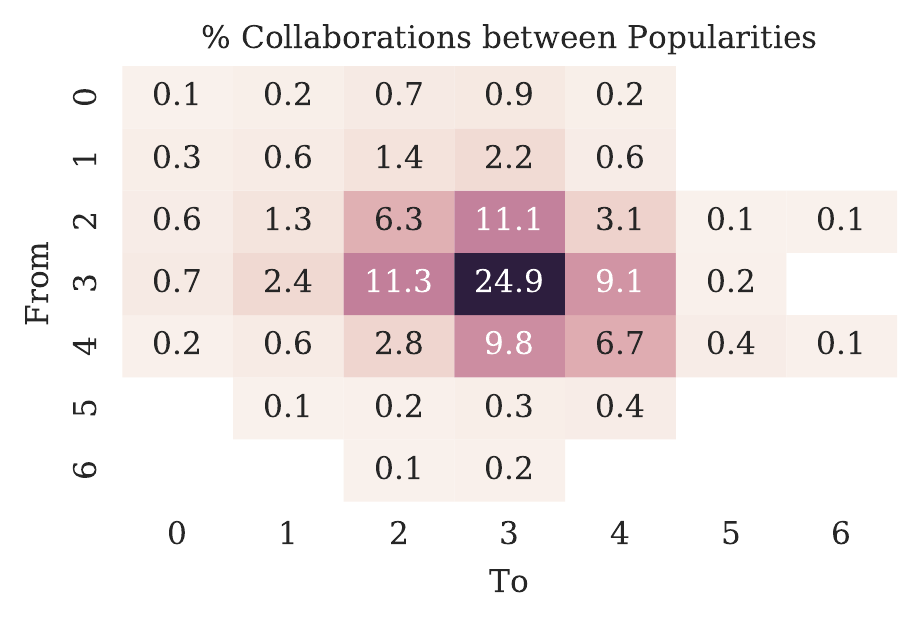}\label{fig:ev_collab_popularity_map}}
	%\hfil
	\subfloat[Video categories.]{\includegraphics[width=0.65\linewidth]{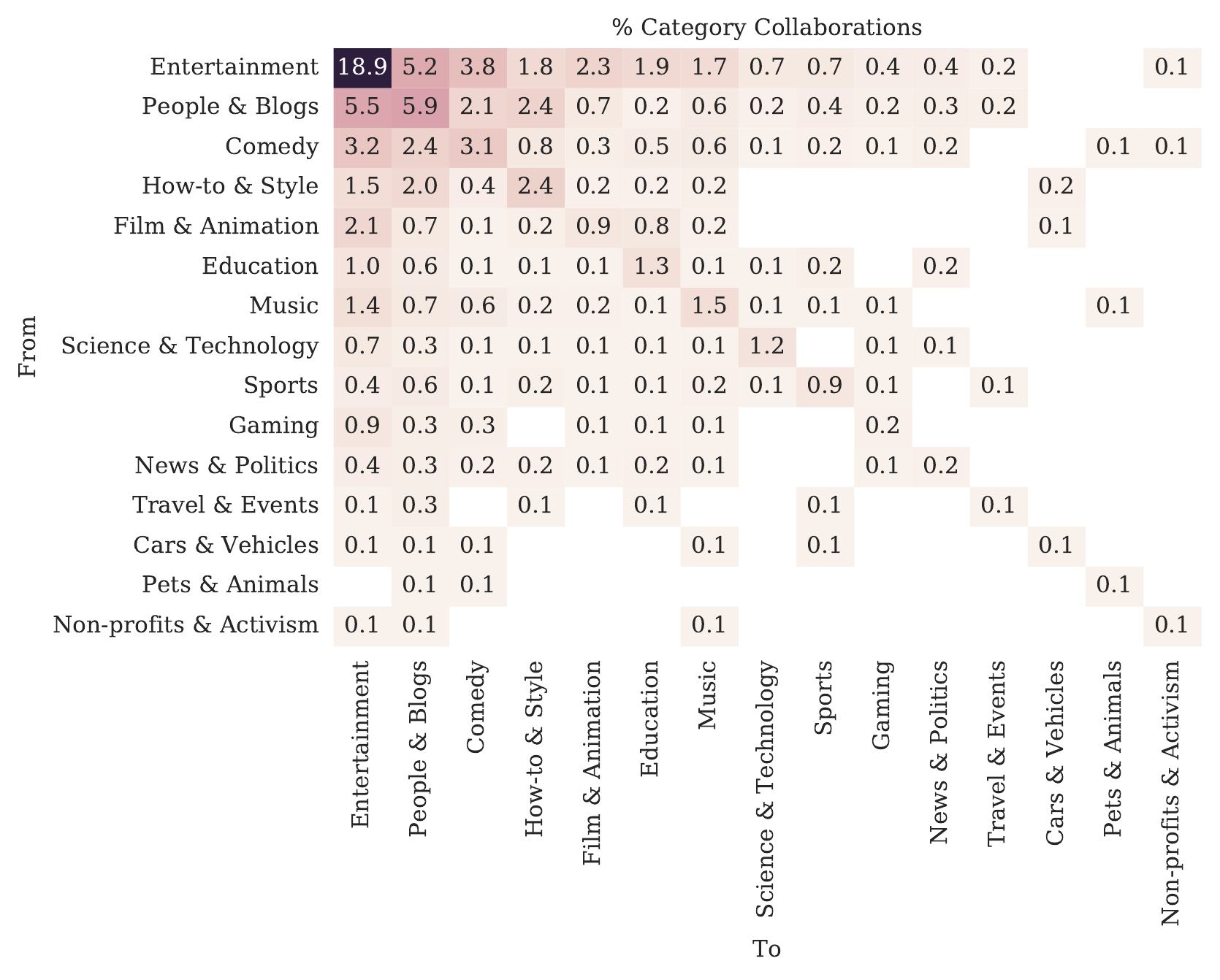}\label{fig:ev_collab_category_map}}
	\caption{Collaborations within \& between popularity classes and YouTube video categories (in \%).\label{fig:collab_in_out_category_popularity}}
\end{figure}

\begin{table}[t]
    \footnotesize
	\centering
	\caption{Popularity class definitions and their number of observed channels.\label{tab:ev_popularity_classes}}
	\begin{tabular}{l l l}
		\toprule
		\textbf{Popularity Class} & \textbf{Subscriber Range}  & \textbf{\#Channels} \\ \midrule
		0                         & $[0, 10^3)$    & 813                        \\
		1                         & $[10^3, 10^4)$  & 1,575                       \\
		2                         & $[10^4, 10^5)$  & 2,569                       \\
		3                         & $[10^5, 10^6)$  & 2,420                       \\
		4                         & $[10^6, 10^7)$  & 544                        \\
		5                         & $[10^7, 5x10^7)$ & 20                         \\
		6                         & $[5x10^7, 10^8)$ & 1                          \\ \bottomrule
	\end{tabular}
\end{table}
In the following, we group YouTube channels with respect to popularity and content category.
First, we assign each channel to one out of seven popularity classes, based on their subscriber count.
Here, the chosen classes resemble the classes used for the YouTube awards\footnote{\url{https://www.youtube.com/yt/creators/rewards.html} [Accessed: \today]}, which are awards shipped to the YouTubers when they exceed a certain number of subscribers.
Next, we analyze collaboration behavior regarding the YouTube video category, a label which the YouTuber can select out of a set of given categories during the video upload process.
%
% Populairty Classes
\paragraph*{YouTube Popularity Classes}
We assign a channel's popularity class with respect to  the number of subscribers as depicted in \Cref{tab:ev_popularity_classes}.
In column \emph{\#Channels}, the table shows also the number of YouTube channels in our dataset which belong to the corresponding popularity class.
We can see that the major share of the channels observed belong to popularity class 1, 2, or 3.
Class 6 is an exception, as it only contains a single channel, i.e., \emph{PewDiePie}, the most successful YouTuber in terms of subscribers so far.
\Cref{fig:ev_collab_popularity_map} depicts the share of observed channel collaborations between popularity classes.
Here, channels belonging to a numerically higher class have more subscribers than channels belonging to numerically smaller classes.
The matrix entry $a_{\text{t} \text{f}}$ of row $f$ and column $t$ denotes that the number of YouTubers \emph{from} a channel of popularity class $f$ appear in videos belonging to channels of popularity class $t$.
We can see that most collaborations happen within class 3 and neighboring classes 2 and 4, which we
ascribe to two factors.
First, these channels have reached a popularity in the YouTube environment that attracts collaborations.
Second, these channels do not yet belong to the most popular channels, i.e., categories 4, 5, and 6 but are likely to try to increase their own popularity by attracting more viewers through collaborations with other YouTubers.
%Examining the plot further, we notice a general trend for collaborations to occur around class 3 and their nearest neighbor classes 2 and 4.
%Additionally, very low popularity classes seem not to collaborate with channels in other classes having a distance of more than one class.
%\emph{PewDiePie} appearances in class 2 channels \textbf{two times} but also in classes 3 and 4.
%Surprisingly, no collaborations with the nearest popularity class 5 were detected.

%
% YouTube Categories
%%%%%%%%%%%%%%%%%%%%%%%%%%%%%%%%%%%%%%%%%%%%%%%%%%%%%%%%%%%%%%%%%%%%%%%%
\paragraph*{YouTube Categories}
In \Cref{fig:ev_collab_category_map} we show the share of collaborations between YouTube categories.
Most collaborations are detected in, and between the \emph{Entertainment} and \emph{People \& Blogs} categories, which is reasonable as they prevalently contain human presence and interaction.
The same applies for categories like \emph{Comedy}.
In the figure, we observe asymmetric relations, e.g., between \emph{Comedy} and \emph{Film \& Animation}, that collaborate more in with \emph{Entertainment} channels than within the same category.
%, which count 58, compared to only 43 collaborations from \emph{Comedy} to \emph{Entertainment}.
For collaborations within a category, i.e., the diagonal of the heat map, we only notice a surge for \emph{Entertainment}.
We observe that the most frequent collaborations occurred within the category \emph{Entertainment}, which also shows the second most video uploads.
Note that \emph{Entertainment} is a rather generic term and can therefore, depending on the YouTuber's interpretation, also include comedy, film, and animation related content.
%This circumstance could further explain the higher number of collaborations with \emph{Entertainment} and other categories.

%=============================================================================
\subsection{How do collaborations impact video and channel popularity?}
\label{subsec:collab_impact_on_popularity}
%=============================================================================
We investigate the two parts of this question separately, focusing, first, on the observed effects on video popularity and, second, on channel popularity.
Therefore, we use popularity statistics of a 3-months period for roughly \textbf{$10^5$} videos, considering videos for which we have at least 12 daily popularity measurements, i.e,  videos being older than 12 days.
We chose 12 days as we observed that most older videos do not receive much more views.
%To investigate popularity time series of 12 days, we filter videos younger than 12 days, resulting in 98,688 videos (80.6\%).
%We achieve comparable results by using only the popularity metrics measured in the first 12 days after the respective video's upload.
%As we want to compare the video popularity of collaboration and non-collaboration videos, channels with no collaborations are not useful for our analysis and thus discarded.
Using CATANA, we deduce a collaboration graph with roughly  $10^4$ edges indicating collaborations within the observed $8\times 10^3$ channels.
%We concentrate on collaborating channels and analyze the traces using CATANA and result in a collaboration graph with 1,214 edges, i.e., collaborations between 1,059 of the 7,942 channels.
%This shows, that features channels do not necessarily collaborate.
%\textbf{@Moritz: Vorher wurden doch nur Kan??le ausgew??hlt auf denen Kolaborationen passiert sind. Wieso existieren dann nicht f??r alle 7.942 Kan??le Kollaborations-Kanten im Graph? Moritz: Nein, die 7942 Kan??le wurden aufgrund der acquaintanceship ausgew??hlt, letztendlich haben davon nur 1,059 kollaboriert.}
%This reduces the data to the final set of 29,284 videos containing 2,406 collaborations.
Using two sets for collaboration and non-collaboration videos, we analyze the maximum values of view and subscriber counts of the 12-days time-span and their gradient.
The benefit of these gradients is that they are not biased by absolute numbers but represent the relative popularity growth.

\subsubsection{Video Popularity}
\begin{figure}[t]
	\centering
	\includegraphics[width=0.95\linewidth]{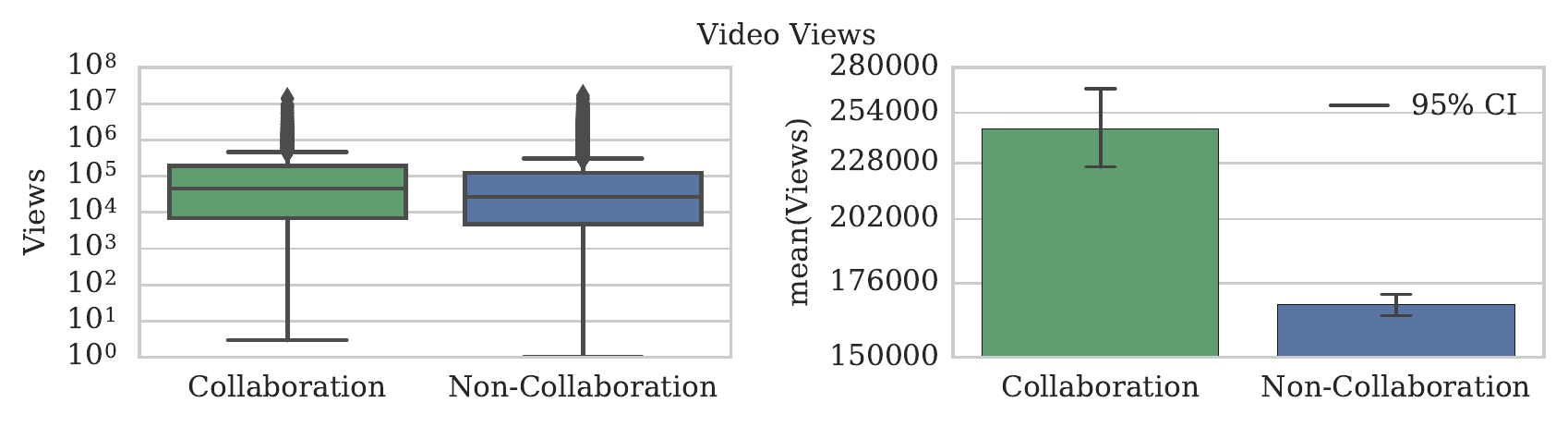}
	\caption{Number of views after 12 days for collab. (green) and non-collab. (blue) videos.}
	\label{fig:ev_collab_video_12day_views}
\end{figure}

First, we examine the maximum video view counts observed in the first 12 days.
\Cref{fig:ev_collab_video_12day_views} shows the video view count distribution. % of the 12 days window.
By examining the left side of the figure, we see in the box plot that the average view count is higher for videos with a collaboration compared with non-collaborations.
Note that the median is stretched nearly doubled (from about 26k to 45k views) for the cases of present collaborations.
%Note that the 50\%-percentile is stretched from about 26k to 45k views in case of a collaboration.
%The difference of the box plot's whiskers is even higher.
Although the median is only slightly higher in case of a collaboration, the upper 50\% of video views are more scattered and show more views.
%\textbf{@Christian: 50\% percentile unterschied ist bei 45k zu 26k views, genauer 18850 views unterschied.}
Additionally, we plotted the average view counts for both cases and their 95\% confidence intervals on the right side of \Cref{fig:ev_collab_video_12day_views} where we see a large gap.
% between the averages.
% of about 77,428 views.
The figure suggests 
%Furthermore, the confidence intervals are not overlapping, suggesting 
that a significantly higher view count can be expected if a video contains a collaboration.
%Though, it becomes clear that the benefits of a collaboration apply only for about half of the observed collaborations resulting in close medians but highly different average view counts.

%\begin{itemize}
%\item \textbf{ToDo: In a next step, we investigate the influence of category, popularity class and repeated collaborations on the effect of view count increase by collaboration}
%\item Hier ware auch interessant weiter zu analysieren, ob es eigenschaften gibt die zu einer erfolgreichen kollaboration gehoeren muessen: Kategorien, Popularitaeten, Kollaborationswiederholungen
%\end{itemize}

%\begin{figure}[t]
%	\centering
%	\includegraphics[width=1.0\linewidth]{figures/collab_video_growth_factor_combined.pdf}
%	\caption{Video view growth factor per popularity and category.}
%	\label{fig:ev_collab_video_views_increase}
%\end{figure}

\begin{figure}[b]
    \footnotesize
	\centering
	\includegraphics[width=0.85\linewidth]{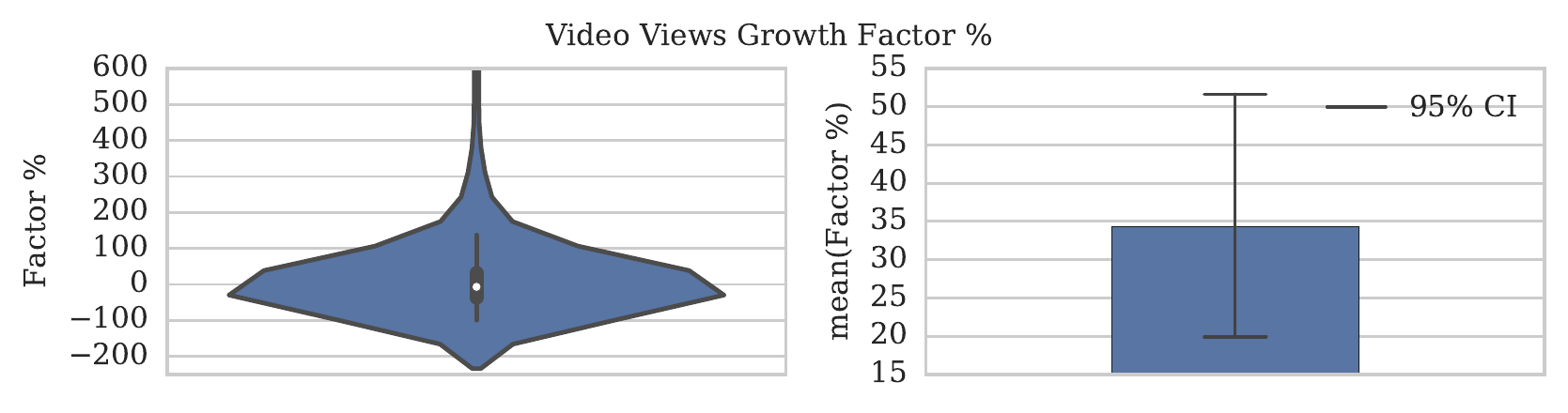}
	\caption{View count growth factor \emph{between} collaboration and non-collaborations videos. A positive value indicates the superiority of collaborations.}
	\label{fig:ev_collab_video_12day_views_growth_factor}
\end{figure}

The generally higher popularity of collaboration videos can be reasoned in the generally higher popularity of collaborating channels.
As we previously examined, class 3 channels collaborate more often than channels of lower popularity classes.
If we therefore assume that more videos with collaborations are uploaded by class 3 channels, the videos consequently tend to a higher popularity compared for example with class 2 videos.
To further substantiate this effect, we evaluate the gradient and growth of the video view counts.
Hence, we calculate the average view growth factor between non-collaboration and collaboration videos.
For this, we use the average maximal 12 day view value for every channel, differentiated by collaboration and non-collaboration.
Next, the percentaged growth from the non-collaboration value to the collaboration value is calculated channel-wise.
\Cref{fig:ev_collab_video_12day_views_growth_factor} displays these results.
As only channel with both collaboration and non-collaboration data could be used for the percentaged growth calculation, samples of 1,116 channels are used. The difference in number of channels is due to channel which only collaborate in external videos, and do not host collaboration in their own.
Statistics on the differences between the two video groups on the video view growth are shown in \cref{table_views_growth}.
\begin{table}[t]
\center
\small
\caption{View growth of collaborations compared to non-collaborations.}
\label{table_views_growth}
\begin{tabular}{ccccccl}
\toprule
Channels & Duration & Mean & Median & 75-percentile & Min    & Max      \\ \midrule
1116      & 3 months & 34.32 & -6.73   & 32.47         & -99.70 & 6,376.28 \\ \bottomrule
\end{tabular}
\end{table}
In a next step, we investigate the distributions of these view count differences depicted in \Cref{fig:ev_collab_video_12day_views_growth_factor}.
Here, the 0.95 confidence interval is between 19\% and 51\%, indicating that a significant growth of the views between collaboration and non-collaboration videos can be expected.
%\textbf{@Moritz: Hier wird nicht 100\% klar, was gegen was verglichen wurde um den Growth Factor zu berechnen, sind es nur Channel-Paarweise vergleiche oder alle Channels durchmischt? Moritz: Es wurden Videos, nicht channel, verglichen. Es wurde pro channel dessen Videos unterteilt in collab und non\_collab, und diese zwei Gruppen innerhalb des channels verglichen, um einen factor zu erhalten um wie viel sich die collab videos innerhalb eines channels unterscheiden.}

\begin{figure}[t]
	\centering
	\includegraphics[width=0.8\linewidth]{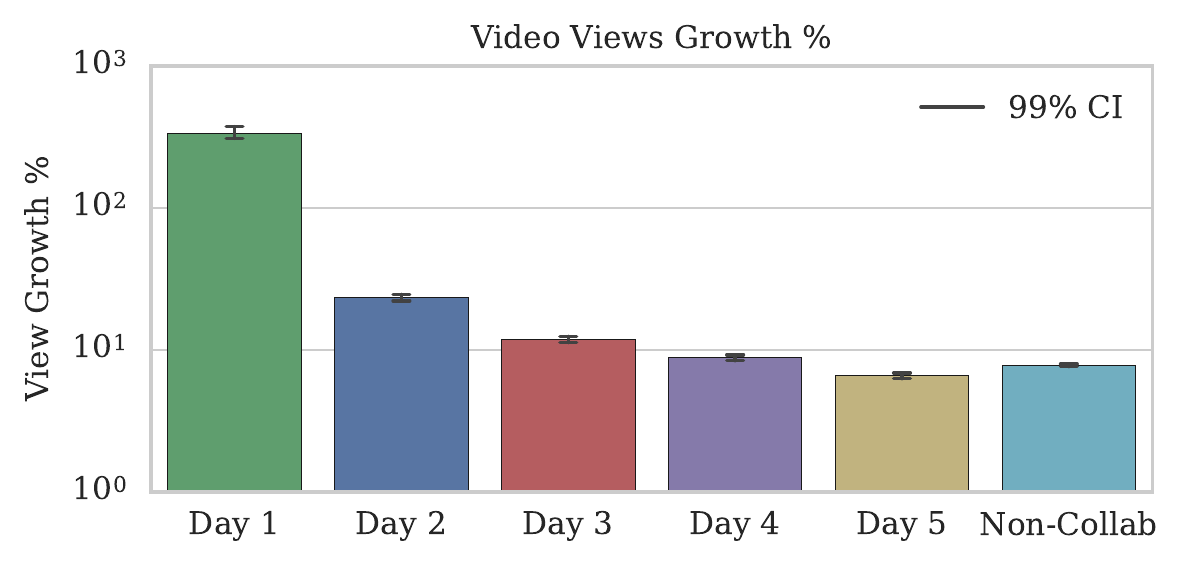}
	\caption{Video view count growth for the first five days of a collaboration vs. the non-collaboration view growth on the first day.}
	\label{fig:ev_collab_video_6day_views_growth_factor}
\end{figure}

\begin{figure}[t]
	\centering
	\includegraphics[width=0.75\linewidth]{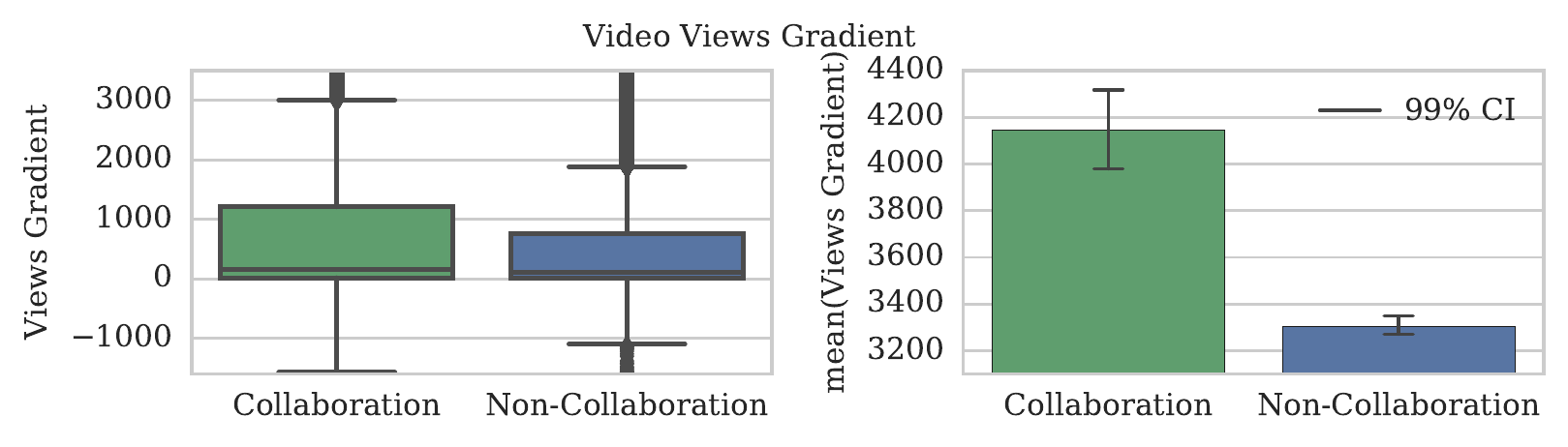}
	\caption{Video view count gradient.}
	\label{fig:ev_collab_video_views_gradient}
    \vspace{0.75cm}
\end{figure}

Next, we compute the gradients between each pair of the 12 days of the videos' view counts, resulting in 11 gradient values.
Additionally, we calculate the percentaged growth between these values.
%, which aims to make the data more independent against heterogeneous channel popularities and, hence, view counts.
The percentaged growth of the first 6 days after a collaboration vs. non-collaboration view growth on the first day is depicted by \Cref{fig:ev_collab_video_6day_views_growth_factor}.
This figure shows the longer lasting temporal impact of collaborations.
%As the view counts are only available on a daily basis, the first day's data in the figure corresponds to the view count growth seen on the second day compared with the first day.
%We can see that till the fourth day, the view count growth is higher than the view count growth of a video after two days.
%
Further, \Cref{fig:ev_collab_video_views_gradient} shows the box and bar plots of the gradient values, i.e. the absolute view count increase, using a 0.99 confidence level.
%The intervals do not overlap, which indicates that the mean view gradient of collaboration and non-collaboration videos differ significantly in favor for collaborations, i.e., on average more than 838.

%=================================
\subsubsection{Channel Popularity}
\begin{figure}[t]
	\centering
	\includegraphics[width=1.0\linewidth]{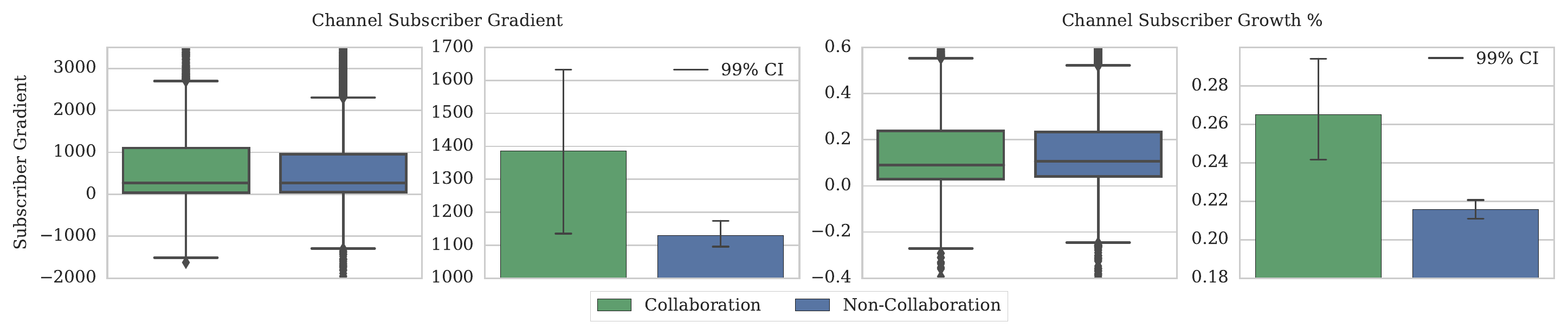}
	\caption{Channel subscriber gradient and percentaged growth.}
	\label{fig:ev_collab_channel_subscriber}
    \vspace{0.75cm}
\end{figure}

%=================================
Here, we cannot directly differentiate between collaboration and non-collaboration channels as for videos, since one channel may contain both, collaboration and non-collaboration videos.
Therefore, we filter channels, which uploaded videos with and without collaborations, resulting in 1,599 channels.
%For these channel we have 67,776 statistic values which can be analyzed.
%
The channel popularity is measured by the number of views of its videos and the number of channel subscribers.
We define a window of two days after a collaboration, for which we will classify the subscriber counts as belonging to a collaboration, the remaining statistics measured are classified as belonging to non-collaboration.
Thereby, we gathered 9,086 channel subscriber measurements
%\textbf{@Moritz: da stand mal statistic values, aber ich bin nicht sicher, ob du damit nich noch andere werte meintest wie die summe an video aufrufen des Kanals, bitte kl??ren. Moritz: Fuer channel war subscriber count und view-count gemeint.}
for collaborations, and 78,877 for non-collaborations.
%
%
%As mentioned in section~\ref{sec:eval_metrics}, an error in the view count data occurred, which we then removed using interpolation.
%As the interpolation of multiple data points may reflect in the evaluation, we additionally leverage the video view count to model a channel's daily view count. Therefore, both the channel view count and the video-based channel view count are evaluated.
%In figure~\ref{fig:ev_collab_bug_sample} we plotted a flawed view count trend from a sample channel, and next to it the video view count induced daily channel view count of the same channel.
\Cref{fig:ev_collab_channel_subscriber} shows the subscriber count differentiated between collaborations and non-collaborations.
%On the figure's left side, we see that the subscriber count gain distribution is increased for about half of the collaboration videos, though, showing a similar mean.
%On the figure's right side, we can see that the mean gain of subscribers are significantly higher in case of a collaboration as the  associated 99\% confidence intervals are not overlapping.
From the figure we conclude that collaborations have a slight positive effect on channel subscribers.
Compared to the increase of viewers, the increase of subscribers is only about one tenth of total users.
Though, the relative growth compared to non-collaborating video measures is about 30\% larger in terms of newly attracted subscribers compared with the newly attracted viewers.

\begin{figure}[t]
	\centering
	\subfloat[Channel subscriber count growth.]{\includegraphics[width=0.49\linewidth]{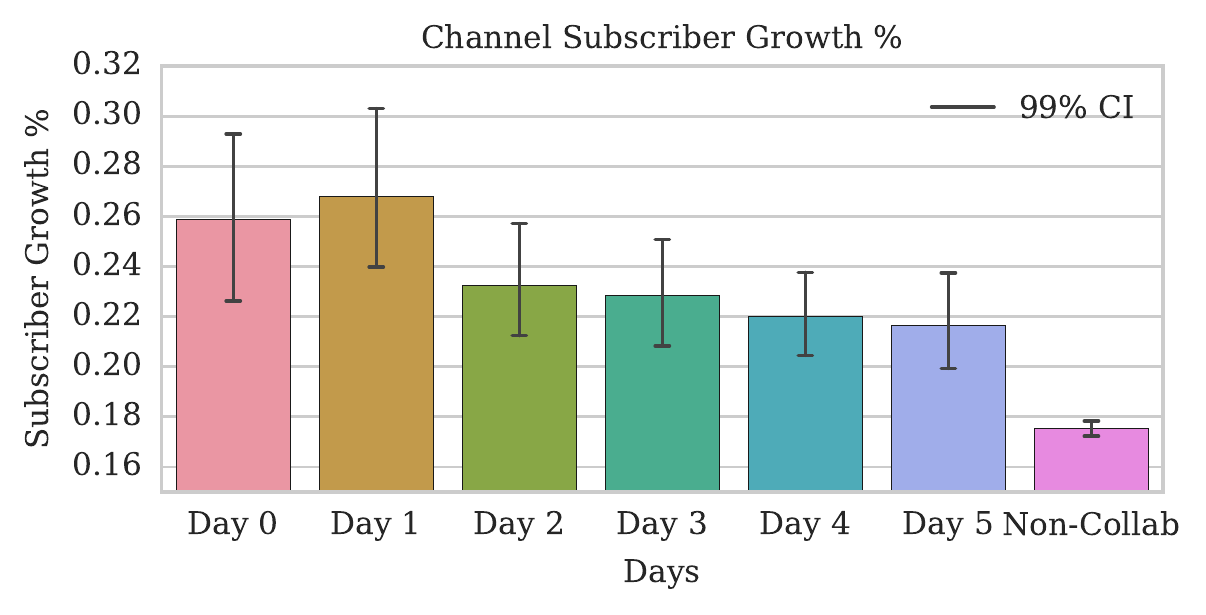}\label{fig:ev_collab_subscriber_course_6day}}
	\hfil
	\subfloat[Channel view count growth.]{\includegraphics[width=0.49\linewidth]{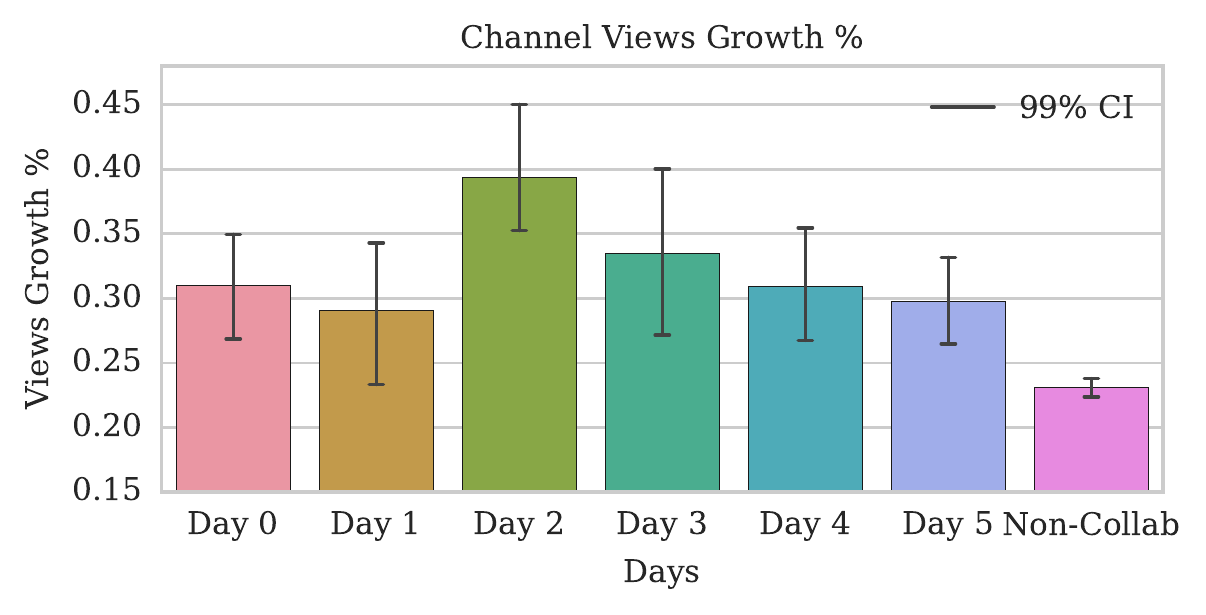}\label{fig:ev_collab_view_course_6day}}
	\caption{Effect of collaboration over the first six days.\label{fig:ev_collab_sample}}
\end{figure}

In addition to the above evaluation of the two-day collaboration window, we evaluate a 6-day window on a daily basis.
\Cref{fig:ev_collab_subscriber_course_6day} shows the percentaged subscriber growth over 6 days starting with day 0, which is the upload date of the collaboration video.
\Cref{fig:ev_collab_view_course_6day} depicts the percentaged subscriber growth for each day and on the left side and the respective overall channel view growth on the right side. 
On both sides, also the growth of non-collaboration videos after day 0 is shown to allow for a comparison.
We note that the highest subscriber growth can be observed for days 0 and 1 with the growth slowly decreasing to approach the base line describing non-collaborations.
In addition to the channel subscriber growth, we apply the same evaluation methodology for the overall channel views, an alternative measure of channel popularity.
Here, we expect a similar pattern as for the channel subscriber counts but we observe a very different pattern, depicted in \Cref{fig:ev_collab_view_course_6day}.
For day zero and one, the view growth is quite low and close to non-collaboration videos.
In contrast to that, for day 2, we observe a significant increase.
%At the time of writing, we have no explanation on why the surge occurs only 2 days after the collaboration.
% we have not been able to detect.
% in the time of this thesis, as due to time constraints
%no additional evaluation concerning this effect could be carried out.
%One explanation for this observation is that on day 2, new users found the videos, or were guided to the videos through the YouTube recommendation, possibly based on wherever they watched, or subscribed one of the collaborating channel.
Despite the fact that more people watch collaboration videos on day two, we found that users seem to be less engaged as for day 2 overall less subscriptions are observed than for days 0 and 1.

%%%%%%%%%%%%%%%%%%%%%%%%%%%%%%%%%%%%%%%%%%%%%%%%%%%%%%%%%%%%%%%%%%%%%%%%
\begin{figure}[t]
	\centering
	\includegraphics[width=1.0\linewidth]{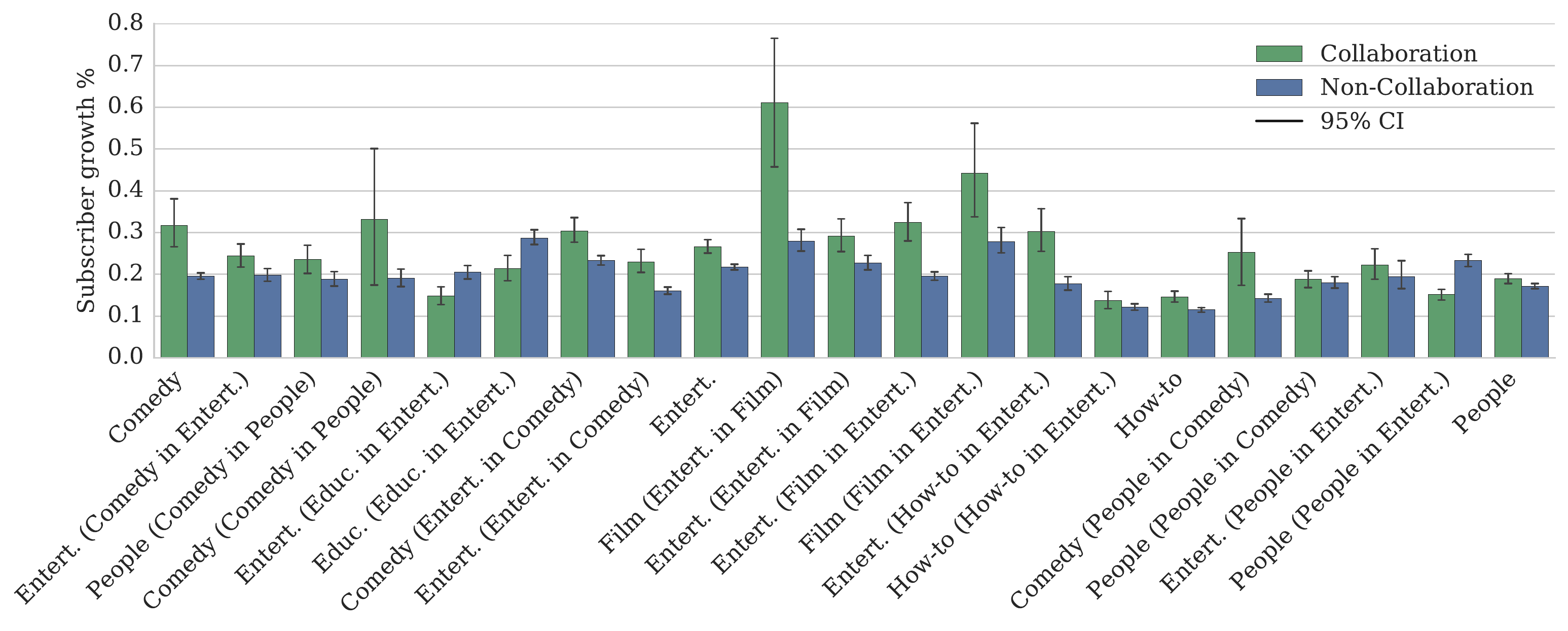}
	\caption{Subscriber growth for collaborations between YouTube channel categories.}
	\label{fig:ev_collab_category_comparision_viewcount}
    
\end{figure}

\begin{figure}[h]
	\centering
	\includegraphics[width=1.0\linewidth]{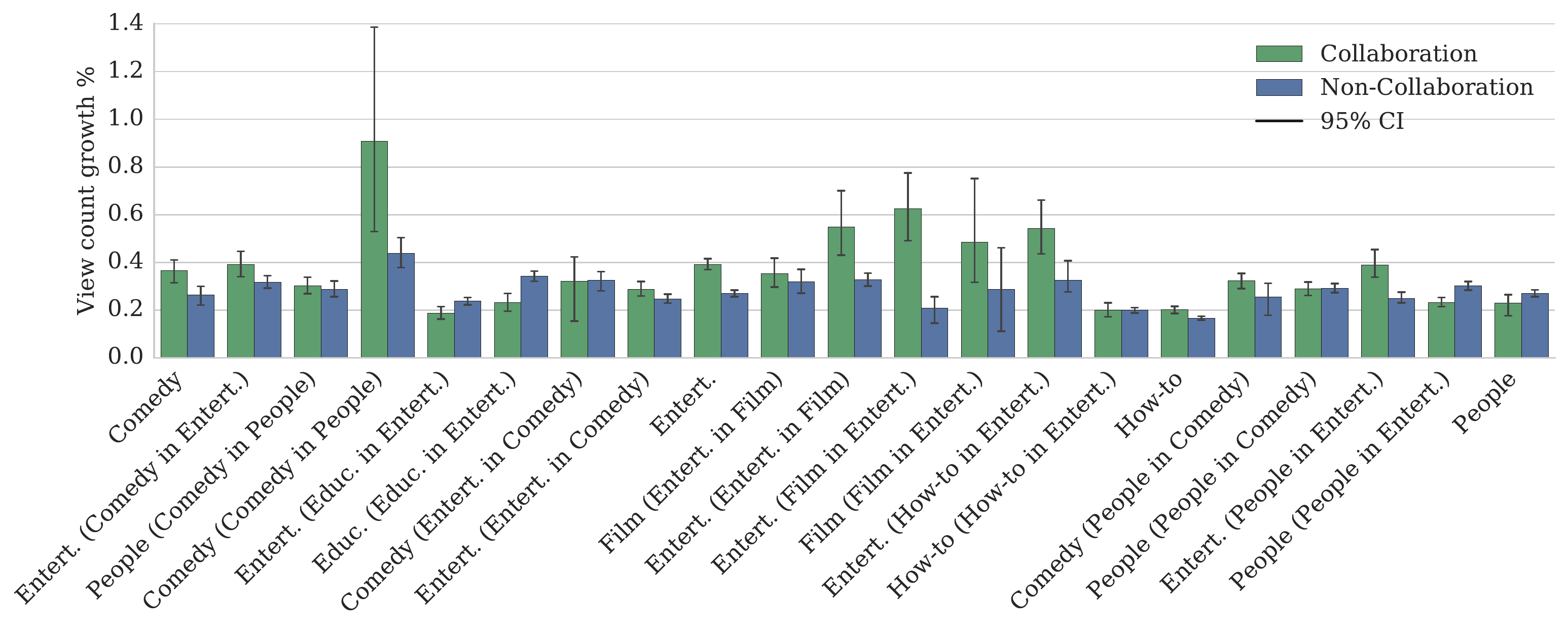}
	\caption{View count growth for collaborations between YouTube channel categories.}
	\label{fig:ev_collab_category_comparision_subscriber}
    \vspace{0.75cm}

\end{figure}
%%%%%%%%%%%%%%%%%%%%%%%%%%%%%%%%%%%%%%%%%%%%%%%%%%%%%%%%%%%%%%%%%%%%%%%%

%%%%%%%%%%%%%%%%%%%%%%%%%%%%%%%%%%%%%%%%%%%%%%%%%%%%%%%%%%%%%%%%%%%%%%%%
\begin{figure}[h]
	\centering
	\includegraphics[width=1.0\linewidth]{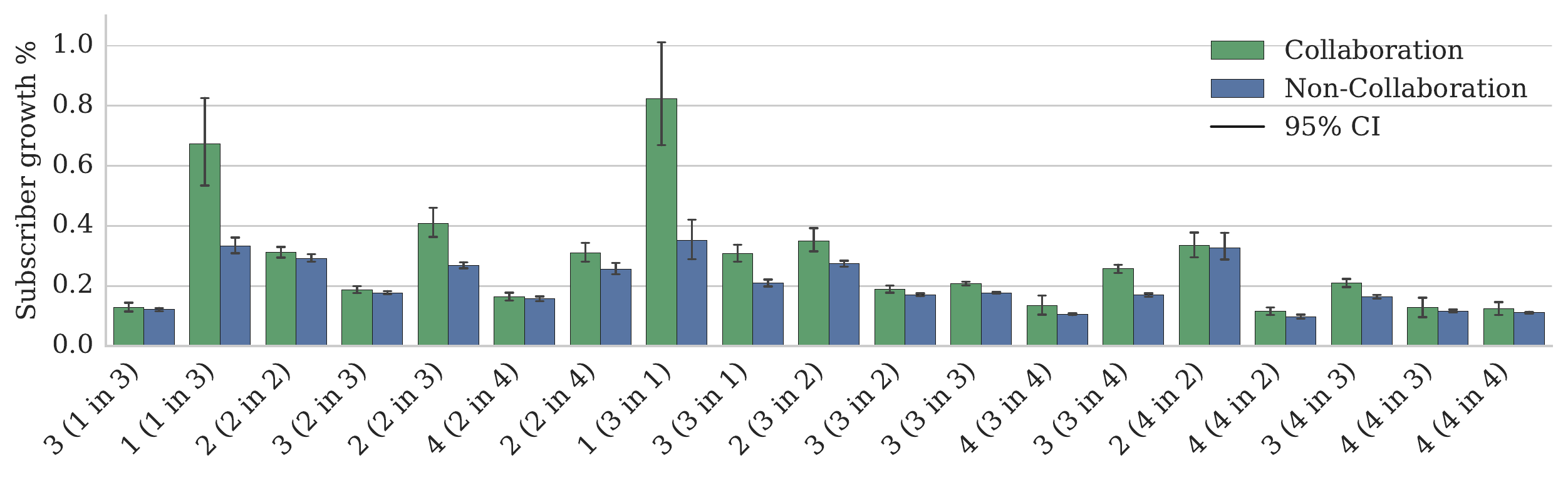}
	\caption{Subscriber growth for collaborations between channel popularity classes. The x-axis denotes the depicted popularity class in the constellation depicted in brackets, e.g., (1 in 3) means that a YouTuber of popularity class 1 appeared in a video published on a popularity class 3 channel.}
	\label{fig:ev_collab_popularity_comparision_viewcount}
    \vspace{0.3cm}
\end{figure}

\begin{figure}[h]
	\centering
	\includegraphics[width=1.0\linewidth]{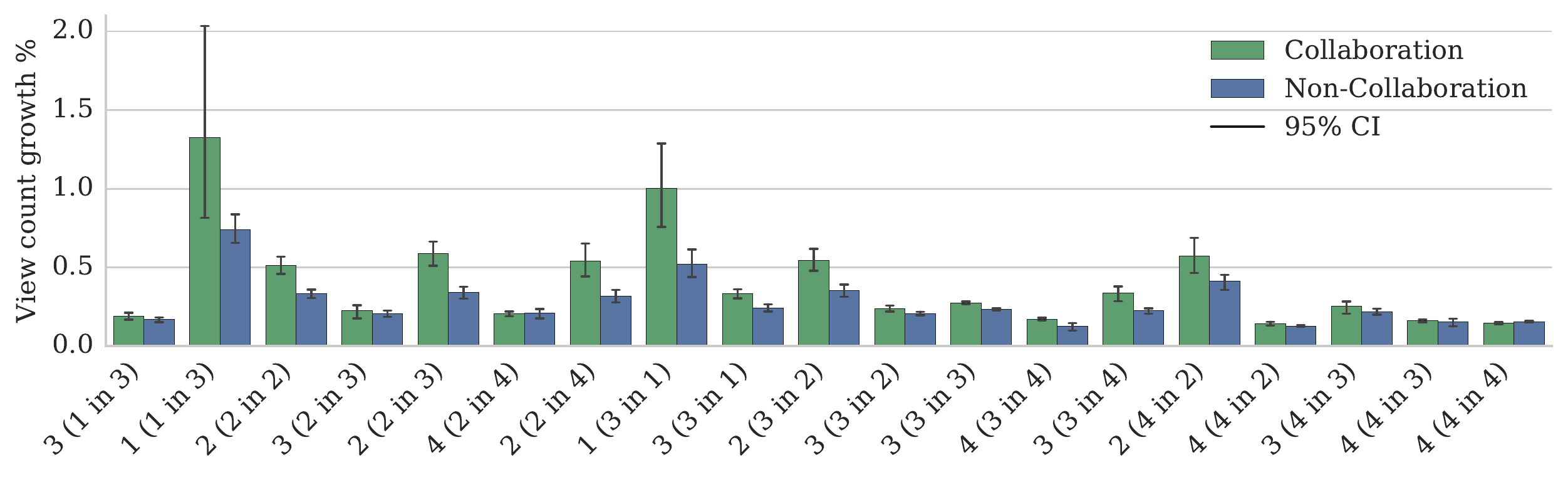}
	\caption{View count growth for collaborations between YouTube channel popularity classes.}
	\label{fig:ev_collab_popularity_comparision_subscriber}
    \vspace{0.3cm}
\end{figure}
%%%%%%%%%%%%%%%%%%%%%%%%%%%%%%%%%%%%%%%%%%%%%%%%%%%%%%%%%%%%%%%%%%%%%%%%

\subsection{Impact of Video Popularity Classes and Video Categories}

%%%%%%%%%%%%%%%%%%%%%%%%%%%%%%%%%%%%%%%%%%%%%%%%%%%%%%%%%%%%%%%%%%%%%%%%
\paragraph{YouTube Categories}
Figure~\ref{fig:ev_collab_category_comparision_viewcount} and \ref{fig:ev_collab_category_comparision_subscriber} show the impact  of a collaboration of YouTubers belonging to different video categories and popularity classes.
We observe that view and subscriber counts vary strongly amongst different channels.
Hence, we compute the relative popularity growth for videos with and without collaborations.
Here, we took only popularity measurements if a collaboration video was uploaded and at the same day no other video was uploaded to guarantee that the channel popularity measure is not impaired.
In case different categories collaborate, we show the effects for both categories separately.
For all collaborations taking place between YouTubers uploading mostly videos of the same category, a significant increase of subscriber and view count is observed.
Here, the category \emph{People \& Blogs} is an exception, as on average more subscribers can be attracted by a collaboration but less video viewers.
This is still beneficial as the subscribers are potentially watching all videos uploaded by the YouTubers in the future.
\vspace{0.5cm}

\paragraph{YouTube Popularity Classes}
In Figure~\ref{fig:ev_collab_popularity_comparision_viewcount} and \ref{fig:ev_collab_popularity_comparision_subscriber} the impact of collaborations between channels of different popularity classes is shown.
In general, we observe a significant benefit of collaborations.
Note that lower popularity classes, especially class 1 and 2 YouTubers benefit significantly stronger than higher popularity classes.
It can be seen that for classes 3 and 4, the gain of a collaboration is comparably low and often there is no significant difference.
For class 4 YouTubers, no effect of collaborations can be observed.
Hence, we deduce that especially for YouTubers with less than $10^5$ subscribers, i.e., class 1 or 2, collaborations significantly increase the number of views.
% and, even more valuable, collaborations.
% Den letzten Teil habe ich nicht verstanden.. es steht nirgends wie collaborations increase collaborations?!
\vspace{0.5cm}

Concluding our popularity evaluation for videos and channels under collaborations, we state we observe that collaborations have positive effects on video views and on engaging new subscribers.
Especially for YouTubers of low popularity classes such as class 1, a collaboration can add about 100\% additional views and new channel subscribers.
Concerning the impact on video views, we calculated a percentaged growth through collaborations with a mean between 19\% and 51\% with 0.99 confidence.
Also channels popularity is significantly increased through collaborations, i.e., for both considered metrics, namely channel subscribers and total channel views.
% Schwach.... aber was tun?... Mehr insights? Selectiver DeepDive?
\vspace{0.5cm}

%====================================
\section{Conclusion and Future Work}\label{conclusion}
%====================================
%Summary	
In this article, we first designed and implemented a system for the acquisition and analysis of collaboration data in user generated content at the example of YouTube.
We implemented a video-based face recognition system named CATANA for which we examined and evaluated different face recognition and clustering techniques.
%We applied CATANA to a set of 81k videos, excluding Gaming videos leaving us with 46k videos.
%We decided to exclude Gaming videos as they often show faces from Game covers and virtual gaming figures as well as often not show the YouTuber but only his gameplay.
We applied CATANA to a collected dataset of videos over a 3-month period where we extracted appearing content creators and leveraged this information to detect collaborations between channels.
%We derived a collaboration graph modeling the collaborations as edges between channel nodes.
%In the process issues regarding embedded content of other channels occurred and were resolved by discarding affected videos and using an additional threshold in the content creator assignment.
%The emerged graph was then leveraged for evaluation and analysis concerning our research questions.

%Conclusion
We observed that out of 7,492 channels in our dataset 1,599 collaborated, with an average of 2.8 times per channel.
%We noticed that collaborating channel pairs tend to collaborate on average of 2.3 times within our evaluated 3 month time-span.
Regarding the types	of channels, which collaborate we found that channels with a subscriber count between \textbf{$10^5$ - $10^6$} collaborate the most and the \emph{Entertainment} YouTube category shows most collaborations.
%, either within or together with other channels of other categories.
%A reason for the dominance in collaborations of the \emph{Entertainment} category is thereby the face recognition approach used to detect the collaborations, as in Entertainment generally a high human presence is examined as compared to, for example, the category "Gaming".
Furthermore, we inferred that multi-channel networks generally exhibit collaborations within the same network or channels, which are not associated with any other, potentially competing, network.
We analyzed the acquired popularity statistics for both videos and channels and found significant differences between collaboration and non-collaboration sets, indicating a positive effect on the popularity that is measured by subscriber and view counts.
In this work we proposed a viable method for collaboration detection in user generated content that is based on face recognition. 
A potential limitation of the proposed system is the differentiation with respect to the collaboration context, which can be mapped to different types of face appearances, i.e. posters or content usage.
In future work, using voice recognition in addition to face recognition is likely to increase detection rates as well as differentiate collaboration context. 
%A future research direction is the detection of collaborations in content with no or little human presence.
%% Future Work
%In future work, we plan to replace our face detection and recognition method by a face tracking mechanisms, e.g., a Kanade-Lucas-Tomasi (KLT) feature tracker[34] to ease the face recognition process as the face within a face track can be assumed to be the same in every frame of the track.
%A potential limitation of the proposed system is the differentiation between actual collaborations and other types of face appearances, i.e. posters or content usage.
%In future work, using voice recognition in addition to face recognition is likely to increase detection rates also for gaming videos.

\section*{Reproducibility}
To enable other researchers to use, reproduce, and extend our research, we release our tool chain at \url{https://github.com/christiannkoch/CATANA}.
This contains:
(i) the CATANA framework and evaluation scripts used here,
(ii) the 3-months collaboration graph used in this paper, and
(iii) an interactive web-based visualization of the collaboration graph.

\section*{Acknowledgements}
This work has been funded in parts by the DFG as part of the Collaborative Research Centre 1053 MAKI (C3, B4).

\newpage
% Bibliography
\bibliography{bibliography}
\bibliographystyle{abbrv}

\end{document}